\documentclass[10pt,journal,compsoc]{IEEEtran}
\ifCLASSOPTIONcompsoc
  \usepackage[nocompress]{cite}
\else
  \usepackage{cite}
\fi
\ifCLASSINFOpdf
\else
\fi
\usepackage{times}
\usepackage{soul}
\usepackage{url}
\usepackage[hidelinks]{hyperref}
\usepackage[utf8]{inputenc}
\usepackage[small]{caption}
\usepackage{graphicx}
\usepackage{amsmath}
\usepackage{amsthm}
\usepackage{booktabs}
\usepackage{enumitem}
\usepackage{subfigure}
\usepackage{multirow}
\usepackage{makecell}
\usepackage{amsfonts}
\usepackage{bm}
\usepackage[linesnumbered,ruled,vlined]{algorithm2e}

\usepackage{scalerel}
\usepackage{tikz}
\usetikzlibrary{svg.path}
\definecolor{orcidlogocol}{HTML}{A6CE39}
\tikzset{
  orcidlogo/.pic={
    \fill[orcidlogocol] svg{M256,128c0,70.7-57.3,128-128,128C57.3,256,0,198.7,0,128C0,57.3,57.3,0,128,0C198.7,0,256,57.3,256,128z};
    \fill[white] svg{M86.3,186.2H70.9V79.1h15.4v48.4V186.2z}
                 svg{M108.9,79.1h41.6c39.6,0,57,28.3,57,53.6c0,27.5-21.5,53.6-56.8,53.6h-41.8V79.1z M124.3,172.4h24.5c34.9,0,42.9-26.5,42.9-39.7c0-21.5-13.7-39.7-43.7-39.7h-23.7V172.4z}
                 svg{M88.7,56.8c0,5.5-4.5,10.1-10.1,10.1c-5.6,0-10.1-4.6-10.1-10.1c0-5.6,4.5-10.1,10.1-10.1C84.2,46.7,88.7,51.3,88.7,56.8z};
  }
}

\newcommand\orcidicon[1]{\href{https://orcid.org/#1}{\mbox{\scalerel*{
\begin{tikzpicture}[yscale=-1,transform shape]
\pic{orcidlogo};
\end{tikzpicture}
}{|}}}}

\newcommand{\itbold}[1]{\textbf{\emph{#1}}}
\newcommand{\corpus}{\mathcal{D}}

\newcommand{\negativeset}{\mathcal{N}}

\begin{document}

\title{Attentive Representation Learning with Adversarial Training for Short Text Clustering}

\author{Wei~Zhang \orcidicon{0000-0001-6763-8146},~\IEEEmembership{Member,~IEEE}, Chao~Dong, Jianhua Yin, and~Jianyong~Wang,~\IEEEmembership{Fellow,~IEEE}

% <-this % stops a space
\IEEEcompsocitemizethanks{
\IEEEcompsocthanksitem W. Zhang (corresponding author) and C. Dong are with the School of Computer Science and Technology, East China Normal University, Shanghai 200062, China (e-mail: zhangwei.thu2011@gmail.com; 51174500084@stu.ecnu.edu.cn).
\IEEEcompsocthanksitem J. Yin is with Department of Computer Science and technology, Shandong University, Qingdao 266237, Shandong, China (e-mail: jhyin@sdu.edu.cn).
\IEEEcompsocthanksitem J. Wang is with the Department of Computer Science and Technology, Tsinghua University, Beijing 100086, China, and also with the Jiangsu Collaborative Innovation Center for Language Ability, Jiangsu Normal University, Xuzhou 221009, China (e-mail: jianyong@tsinghua.edu.cn).
%\IEEEcompsocthanksitem This paper was partially supported by the National Key Research and Development Program of China (No. 2020YFA0804503), the National Natural Science Foundation of China (No. 62072182, 61802231, 61532010, 61521002), and Beijing Academy of Artificial Intelligence (BAAI).
}% <-this % stops an unwanted space
%\thanks{Manuscript received XX, XXX; revised XX, XXXX.}

}


\markboth{Journal of IEEE TRANSACTIONS ON DATA ENGINEERING,~Vol.~xx, No.~xx, xxxx}%
{Shell \MakeLowercase{\textit{et al.}}: Bare Demo of IEEEtran.cls for IEEE Journals}



\IEEEtitleabstractindextext{%
\begin{abstract}
Short text clustering has far-reaching effects on semantic analysis, showing its importance for multiple applications such as corpus summarization and information retrieval.
However, it inevitably encounters the severe sparsity of short text representations, making the previous clustering approaches still far from satisfactory.
In this paper, we present a novel attentive representation learning model for shot text clustering, wherein cluster-level attention is proposed to capture the correlations between text representations and cluster representations.
Relying on this, the representation learning and clustering for short texts are seamlessly integrated into a unified model.
To further ensure robust model training for short texts, we apply adversarial training to the unsupervised clustering setting, by injecting perturbations into the cluster representations.
The model parameters and perturbations are optimized alternately through a minimax game.
Extensive experiments on four real-world short text datasets demonstrate the superiority of the proposed model over several strong competitors, verifying that robust adversarial training yields substantial performance gains.
\end{abstract}

\begin{IEEEkeywords}
short text clustering, representation learning, attention mechanisms, robust adversarial training
\end{IEEEkeywords}}

\maketitle

\IEEEdisplaynontitleabstractindextext

% \fi
\IEEEpeerreviewmaketitle

\IEEEraisesectionheading{\section{Introduction}\label{sec:introduction}}

\IEEEPARstart{R}{ecent} years have witnessed the fast-growing trade of short text data in various kinds of social media, for example, Twitter, Instagram, and Sina Weibo.
As a consequence, short text clustering, the task of automatically grouping multiple unlabeled texts into a number of clusters, has become increasingly important.
It can benefit multiple content-centric downstream applications, such as event exploration~\cite{Feng15-ICDE}, trend detection~\cite{MathioudakisK10-SIGMOD}, online user clustering~\cite{LiangYK19}, cluster-based retrieval~\cite{LiuC04,CarpinetoR12}, to name a few.
Compared with general text clustering, short text clustering is more challenging.
This is because text representations in the original lexical space are usually sparse and this issue is further amplified for short texts~\cite{Aggarwal12-Book}.
Thus, the key to the success of short text clustering is to learn an effective short text representation scheme suitable for clustering.

On the basis of classical general clustering algorithms such as K-means~\cite{Jain10-PRL}, current developments in short text clustering mostly fall into two branches: Bayesian topic models (e.g., Latent Dirichlet Allocation (LDA)~\cite{Blei03-JMLR}) and deep learning approaches~\cite{lecun2015deep}.
The former one realizes probabilistic text clustering by assuming that each document is associated with a distribution over topics, and each topic is a distribution over words.
In this way, a topic is usually regarded as a cluster.
To model short text, some elaborate topic models~\cite{Yan13-WWW,Yin14-KDD} are presented by changing the text generation process.
However, one major limitation remains within most of these topic models: the input representation of short text is commonly based on the bag-of-words assumption and one-hot encoding, which might be sparse and lack the expressive ability.

Aiming to leverage the power of representation learning for short text clustering, the studies~\cite{Xu15-NAACL,Xu17-NN} utilize word embeddings~\cite{Mikolov13-NIPS} and deep convolutional neural networks~\cite{Kalchbrenner14-ACL} to build a multi-stage framework.
Better text representations are firstly learned to be fed into the conventional K-means algorithm for improving the clustering performance.
However, the optimization process is partitioned into separate stages.
Thus it is incapable of guiding text representation learning by the clustering objective.
Some other deep representation learning approaches for general clustering problems~\cite{Xie16-ICML,Jiang17-IJCAI,Yu18-IJCAI,Hadifar19-ACL} have been trained in an end-to-end fashion, whereas they are not tailored for textual data or require an additional step to obtain short text representations.
Specifically, hand-crafted text representations, such as Term Frequency-Inverse Document Frequency (TF-IDF), are commonly taken as model input.
Hence they overlook the usage of word-level distributional representations, which are important to grasp the text semantic relatedness.

In this paper, we concentrate on bridging the gap between short text representation learning and short text clustering by fusing them in a unified model.
Inspired by the roaring success of attention mechanisms~\cite{Bahdanau2014neural} in Natural Language Processing (NLP), we devise the novel Attentive Representation Learning (ARL) model tailored for the short text clustering task.
It first leverages low-dimensional word embeddings to build dense representations with the simple but effective mean-pooling technique for short texts.
Cluster-level attention is then developed to capture the correlations between the short text representations and cluster representations, which shares a similar spirit with GSDMM~\cite{Yin14-KDD} tailored for short text modeling that assumes all the words in a short text should belong to the same cluster (topic).
The derived attention weights can be used as an evidence to determine cluster assignments.
To enable the unsupervised learning of word and cluster representations, we propose to reconstruct short text representations through the weighted combination of cluster representations.
An objective function combining the pairwise ranking loss and point-wise loss is employed to measure the reconstruction gap.

To improve the effectiveness of learning short text clustering, we further incorporate robust adversarial training~\cite{Goodfellow15-ICLR,Miyato17-ICLR} into the final model, naming it as ARL-Adv.
Robust adversarial training requires to feed both original real examples and intentional ``adversarial examples'' into the model during the training process.
It has obtained impressive performance in the supervised and semi-supervised classification tasks.
Concretely, ARL-Adv associates continuous cluster representations with adversarial perturbations to form an additional adversarial objective function.
As a supplement to the original objective function, adversarial perturbations play a role of adaptive regularization, providing the optimization process with more robustness and being beneficial for short texts which sometimes contain noise.
The model parameters and perturbations are optimized through a minimax game, where the model parameters are for minimizing the aforementioned reconstruction gap, and by contrast, the perturbations are for maximizing the reconstruction gap.

To sum up, our contributions are as follows:
\begin{itemize}
\item We present a novel attentive representation learning approach (ARL) to couple short text representation learning and clustering in a unified model.
By proposing to reconstruct short-text representations through cluster-level attention, the model can be optimized in an unsupervised manner while cluster assignments are learned as well.

\item To enhance the robustness of learning cluster label assignments for short texts, we introduce robust adversarial training by customizing adversarial examples (ARL-Adv).
To our best knowledge, this is the first study to apply robust adversarial training into the field of the unsupervised clustering setting.

\item We conduct extensive experiments on four real-world short text datasets.
Three of them are publicly available, and the remaining one is constructed following a recent work.
By comparing ARL-Adv with several strong baselines, we demonstrate its significant improvements and verify that the adversarial training further boosts performance.
To ensure the reproducibility of this paper, we make the event-related dataset released for evaluation (see Section~\ref{exp-sf}).
\end{itemize}

\section{Related Work}\label{sec:related}
This section briefly reviews recent achievements on text clustering methods, especially for short texts, and generative adversarial networks for clustering.
Besides, text representation schemes, attention mechanisms, and adversarial training are discussed.

\subsection{Text Clustering Methods}\label{subsec:text-clu}
While general text clustering has been extensively studied in the literature~\cite{LiLC08,Aggarwal12-Book,HuangYWZS13,AilemRN17}, short text clustering~\cite{Banerjee07-SIGIR,Rangrej11-WWW} receives less attention until recent years, due to the flourish of user-generated content in online social media. 
Currently, there are two main categories of approaches in the literature.

The first category is about topic modeling approaches, inheriting the ideas from LDA to uncover latent topic distributions by implicitly modeling word co-occurrence patterns.
Unlike LDA that assumes to generate a topic for each word in a document, GSDMM~\cite{Yin14-KDD} supposes all the words in a short text share the same topic, which is naturally regarded as the cluster of the text.
Another advanced model is BTM~\cite{Yan13-WWW}.
It explicitly models co-occurrence patterns through bigrams defined as pairs of unordered words in a given document. 
The cluster determination in BTM depends on the topic proportions of bigrams in a target short text.
More recently, two topic models~\cite{Liang16-KDD,Yin18-KDD} are developed for clustering streaming short texts incrementally.
In this paper, we primarily concentrate on the general clustering setting and leave how to adapt our model to streaming short texts in the future.

Inspired by the success of deep learning~\cite{Bengio09-Book}, the latter category promotes its development in the clustering field.
The most relevant study is STCC~\cite{Xu15-NAACL,Xu17-NN} designed for short text clustering.
However, it is not trained in an end-to-end fashion, leaving room for performance improvement.
Gaussian-LDA~\cite{Das15-ACL} incorporates word embeddings into the generation process of topic models, but it is not very compatible with short texts, just as LDA.
Recently, some deep learning approaches are proposed for general clustering tasks, but not crafted for texts.
DEC~\cite{Xie16-ICML} utilizes the idea of self-training to measure the gap between the distribution of soft cluster assignments and its corresponding auxiliary distribution through Kullback-Leibler (KL) divergence. 
STC~\cite{Hadifar19-ACL} adopts self-training as well, meanwhile, Smoothed Inverse Frequency (SIF) is introduced to weight word embeddings in each document.
Yet STC obtains text representations in a separate step and cannot optimize word embeddings along with learning text clustering.
VaDE~\cite{Jiang17-IJCAI} extends variational auto-encoder~\cite{Kingma14-ICLR} to integrate a Gaussian mixture model to generate latent vectors, instead of just using a Gaussian distribution.
However, all the above models do not address the importance of representation learning for short texts, making their clustering performance not very satisfied in this situation.

\subsection{Generative Adversarial Networks for Clustering}\label{subsec:adv-clu}
Benefiting from generative adversarial networks (GANs)~\cite{Goodfellow14-NIPS} which aim to generate data that matches a target distribution, several approaches utilize GANs for clustering tasks.
Specifically, Premachandran et al.~\cite{premachandran2016unsupervised} employed GANs to extract features, which are later fed into the K-means++ clustering algorithm.
Therefore, it is not learned in an end-to-end fashion for clustering. 
DAC~\cite{Harchaoui17-ICLR} leverages a discriminator to push forward the distribution of the data representations generated by a trainable encoder to a Gaussian mixture distribution.
GANMM~\cite{Yu18-IJCAI} revisits the expectation-maximization based clustering algorithm and replaces the expectation step and the maximization step with classification models and GANs.
Mrabah et al.~\cite{Mrabah-Arxiv19} extended DEC~\cite{Xie16-ICML} by adding a regularization term through GANs to constrain the reconstructed data points to be similar to realistic data points.
ClusterGAN~\cite{MukherjeeALK19} consists of a generator, a discriminator, and an encoder to learn mappings between discrete class labels and continuous image pixels.
The innovation lies in the design of a discrete latent space $Z$, thereby providing sufficient signals for the generator to have samples belonging to a specific class and further benefiting clustering.
The study~\cite{GhasediWDH19} shares a similar spirit with ClusterGAN by devising a special latent space for ease of clustering.
Moreover, self-paced learning is adopted to distinguish easy and hard examples for achieving a reliable training process.
In summary, most of the GAN-based methods are welcomed for clustering images due to their ability to handle continuous values, whereas it is nontrivial for applying GAN-based methods to clustering the discrete textual data due to the difficulty of optimization~\cite{YuZWY17}.
A surrogate way is to represent text as fixed continuous vectors (e.g., TF-IDF or pre-trained word embeddings) for GAN-based models.
However, it might inevitably limit the power of learning text representations.

\subsection{Text Representation Schemes}\label{subsec:textrep}
Traditionally, a single word is represented by a one-hot vector with a fixed dimensional space.
The space has the same size as that of a specified vocabulary.
Accompanied by this, vector space models with TF-IDF~\cite{manning2008introduction} are commonly utilized to denote short text representations.
To alleviate the issue of sparse exact word overlap in similarity computation, low-dimensional techniques, such as Principal Component Analysis (PCA) and probabilistic topic models (e.g., LDA), are applied to the text domain.
In particular, word embedding techniques~\cite{Mikolov13-NIPS} have gained lots of attention in recent years, due to their improved ability to grasp semantic relatedness.
Consequently, much progress has been observed in this respect, including contextualized word embeddings like BERT~\cite{DevlinCLT19} and GPT-3~\cite{brown2020language}. 
Benefiting from the advantage of word embeddings, the pre-training and fine-tuning paradigm~\cite{xipengpre} becomes the state-of-the-art for text classification and some other NLP tasks.
Nevertheless, the embedding based text clustering approaches~\cite{Xu15-NAACL,Xie16-ICML,Hadifar19-ACL} could not fine-tune the fine-grained word-level representations by the guidance of clustering objectives, thus leaving room for improving text representations.
This motivates the proposed models ARL and ARL-Adv to mitigate this issue by jointly learning word embeddings and text clustering (see the degraded performance of not fine-tuning word embeddings in Table~\ref{tbl:ablation}).

\subsection{Attention Mechanisms}
Attention mechanisms are growing in popularity for learning representations for image, text, and other different modalities~\cite{Bahdanau2014neural,Vaswani17-NIPS,YeungRJAMF18,XiaoYFSYZ19,GaoLSS20}.
For a given context (query), it typically calculates the prominence of each representation (key) in sub-layers to form an integrated representation (value).
In the literature of text modeling, attention mechanisms allow for dependency modeling regardless of the distance between words, compared with recurrent neural networks~\cite{Hochreiter97-NC}.
The recent study~\cite{Vaswani17-NIPS} develops a multi-header self-attention mechanism for representing input and output sequences.
While the approach is entirely composed of attention blocks, it outperforms several recurrence and convolution based models in machine translation tasks.
Based on this self-attention mechanism, large-scale pre-trained language models (e.g., BERT) have been proposed and benefited many NLP tasks.
For the specific text classification task, Ma et al.~\cite{MaYTCN19} introduced a mutual attention mechanism to assign importance weights to both semantic related long-distance words and local semantic features.
In another aspect, the works~\cite{Chen16-EMNLP,Wang18-IJCAI} consider incorporating user factors into the attention computation process for words and sentences.
The work~\cite{he2017unsupervised} also adopts word-level attention to extract aspect-related words and de-emphasize irrelevant words.
Different from these studies, we perform cluster-level attention to associate the representation learning with the automatic determination of cluster assignments in a unified model to achieve the state-of-the-art short text clustering performance.

\subsection{Robust Adversarial Training}\label{subsec:Robust-AT}
Since the pioneering work~\cite{Goodfellow15-ICLR} proposes to investigate the benefit of supervised learning on intentional adversarial examples for image classification, robust adversarial training has made continuous progress in different fields~\cite{Miyato16-ICLR,Miyato17-ICLR,He18-SIGIR,Yan18-NeurIPS}.
For example, Miyato et al.~\cite{Miyato17-ICLR} applied virtual adversarial training~\cite{Miyato16-ICLR}, a variant of adversarial training for semi-supervised learning, to the text classification task.
He et al.~\cite{He18-SIGIR} learned a pairwise ranking based matrix factorization model for achieving better recommendation performance in an adversarial manner.
In a nutshell, the peculiarities of robust adversarial training are to proactively inject small intentional perturbations to the original data and train models to behave well on them, thereby enhancing the robustness of model training.
To the best of our knowledge, this is the first study to leverage robust adversarial training in the literature of unsupervised clustering.

It is worth noting that robust adversarial training has a remarkable difference with GANs due to the following reasons.
First of all, unlike GANs aiming at generating new data to match a specified target distribution, robust adversarial training targets at ensuring a robust model training process.
Secondly, there is no need for robust adversarial training to have a tailored discriminator like GANs since the target is not to distinguish real or synthetic data.
Because of this, some studies~\cite{FuTC19,ZhangGMO19} that couple GANs with attention mechanisms do not share the same technological thought, let alone they are not proposed for clustering tasks. 
Besides, although Wei et al.~\cite{Wei-Arxiv18} adopted adversarial attacks for clustering, their approach is largely different from the proposed ARL-Adv since: (1) It relies on a few labeled data in its clustering procedure. (2) The grid based clustering and adversarial attack are separated in different stages. (3) The adversarial part does not involve model parameter learning.

\begin{figure*}[!t]
	\centering
	\includegraphics[width=0.95\linewidth]{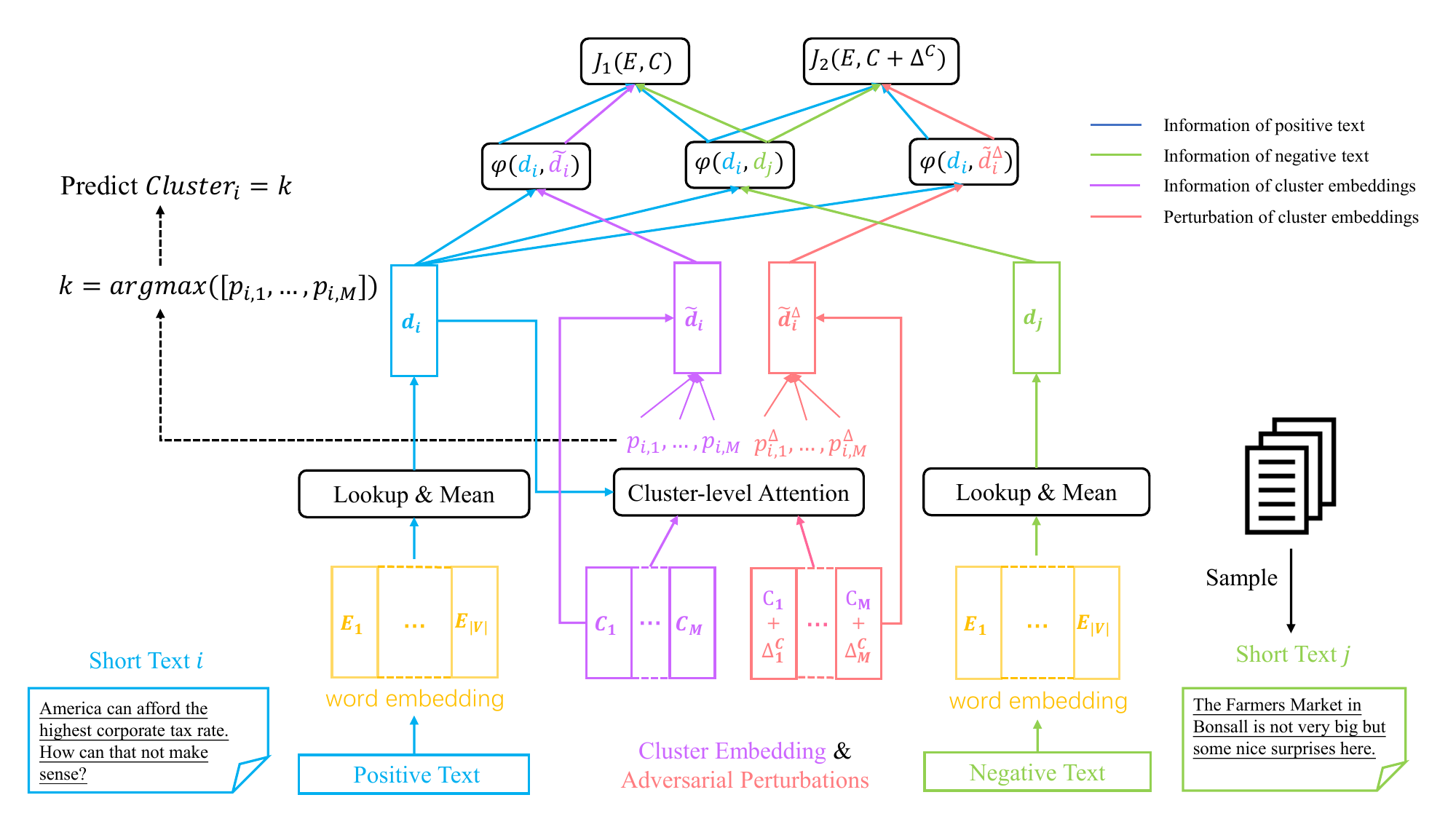}
	\caption{
		Architecture of the proposed ARL-Adv. Four types of marks with different colors are used to label different information flow.
        The dotted lines in the middle left of the figure denote how the cluster label of short text $i$ is derived after the model is well trained. $J_1$ and $J_2$ correspond to the normal objective function (Equation~\eqref{eq:j1}) and the adversarial objective function (Equation~\eqref{eq:j2}), respectively. 
	}
	\label{fig:model}
\end{figure*}

\section{The Computational Model}\label{sec:computation}
\textbf{Model Overview:} 
Figure~\ref{fig:model} depicts the overall architecture of the proposed model ARL-Adv with given short text examples.
From a whole perspective, the model contains five computational modules: (1) short text representation, (2) cluster-level attention, (3) short text reconstruction, (4) objective function, and (5) robust adversarial training.
Taking the given short text (i.e., text $i$) as an illustrative example, it is first fed into the short text representation module (see Section~\ref{subsec:shor-text-rep}), where the words (e.g., tax) in the text are associated with trainable word embeddings to constitute its text representation $\itbold{d}_i$.
Afterwards, cluster-level attention (see Section~\ref{subsec:cluster-level-att}) is conducted to compute the probability (relevance) weights between the text representation and different cluster representations (e.g., $\itbold{c}_1$) that are trainable as well.
The weights are consequently adopted to combine cluster representations for reconstructing short text representation $\widetilde{\itbold{d}}_i$ (see Section~\ref{subsec:text-reconstruct}).
To build the pairwise ranking based clustering objective function (see Section~\ref{subsec:obj-func}), short text $j$ is sampled from the given corpus as a pseudo negative text.
Eventually, the adversarial training module (see Section~\ref{sec:adversarial-train}) injects perturbations (e.g., $\Delta^C_1$) into the cluster representations to constitute another short text reconstruction $\widetilde{\itbold{d}}^\Delta_i$ for ranking, thereby providing the ability of robust training.
After ARL-Adv is well trained, the cluster label of short text $i$ is determined by sorting the corresponding probability weights and returning the cluster index (e.g., $k$) with the maximal weight (e.g., $p_{i,k}$).

\subsection{Short Text Representation}\label{subsec:shor-text-rep}
For the short text clustering problem, we have a corpus $\corpus$ containing multiple short texts, i.e., $\{D_1,D_2,\cdots,D_{|\corpus|}\}$, where $|\corpus|$ is the size of $\corpus$.
Taking the $i$-th short text $D_i\in\corpus$ for illustration, it is associated with a matrix based representation $\itbold{D}_i\!\!=\!\! \{\itbold{x}^i_t\}^{l_i}_{t=1}$, where $l_i$ is the length of the text and $\itbold{x}^i_t$ is the column vector of one-hot encoding at position $t$, with the size equal to the number of words in a pre-specified vocabulary $V$.
It is worth noting that here $\itbold{D}_i$ is the original text representation, while some other strategies such as TF-IDF can be regarded as the hand-crafted feature engineering for texts.

To convert each one-hot sparse representation $\itbold{x}^i_t$ $(t\in\{1,\cdots,l_i\})$ to its low-dimensional dense embedding, we first adopt a look-up encoding procedure:
\begin{equation}
\itbold{w}^i_t=\itbold{E}\itbold{x}^i_t\,,
\end{equation}
where $\itbold{E}\in\mathbb{R}^{K \times |V|}$ is a trainable word embedding matrix and $K$ is the embedding size.
Given these dense word embeddings, a simple mean-pooling technique is empirically adopted to construct an informative representation of the short text:
\begin{equation}\label{eq:dt}
\itbold{d}_i=\frac{1}{l_i}\sum_{t=1}^{l_i}\itbold{w}^i_t\,.
\end{equation}

The motivation behind the above operation is that short texts commonly have insufficient contextual information and might sometimes contain noisy word information.
As such, complex contextual text representation methods, including pre-trained contextual word representations (e.g., BERT), attention based document-level representations~\cite{Yang16-NAACL}, recurrent sequential representations~\cite{Tang15-EMNLP}, and convolution-based representations~\cite{Kim14-EMNLP}, do not exhibit significant improvement and are sometime even worse than mean-pooling in our local tests.
We have also attempted VLAWE~\cite{IonescuB19} for text representations, whereas it is not very suitable because the word embeddings are hard to be fine-tuned when learning short text clustering due to its separate procedure of word embedding clustering.
Consequently, we deem that mean-pooling is suitable for the simultaneous learning of short text representations and clustering.

\subsection{Cluster-level Attention}\label{subsec:cluster-level-att}
To represent the presumed $M$ clusters for the specific corpus $\corpus$, a trainable cluster representation matrix $ \itbold{C}=[\itbold{c}_1,\itbold{c}_2,\cdots,\itbold{c}_M]\in\mathbb{R}^{K \times M} $ is established.
As $\itbold{d}_i$ has contained useful global semantic information for the $i$-th short text, leveraging it to select a suitable cluster can be feasible, which is the primary motivation of cluster-level attention.
In general, an attention function takes a query and a set of key-value pairs as input.
Each attention score characterizes the compatibility of its corresponding key with the query~\cite{Vaswani17-NIPS}, and is later used for the weighted combination of all values.
We let both key and value correspond to the representations of clusters, and regard short text representations as queries.
It is intuitive to perform selection based on a probability distribution $p(z_m|\itbold{d}_i)$ over each cluster $z_m$ where $m\in\{1,2,\cdots,M\}$.
Formally, the distribution is defined as follows:
\begin{align}
z_m \sim p(z_m|\itbold{d}_i) = \frac{\exp(\itbold{c}_m^\top \itbold{d}_i)}{\sum_{m'=1}^M \exp(\itbold{c}_{m'}^\top \itbold{d}_i)}\,.
\label{eq:z_m}
\end{align}

The above computational manner is the realization of cluster-level attention, which captures the text-cluster interaction.
The idea behind it also conforms to the successful experience of probabilistic topic models for short text modeling.
Specifically, in LDA-like models, the topic of each word in a long text is separately determined from their sampled topics.
By contrast, GSDMM assumes that all the words in the same short text belong to the same topic.
It exhibits improved performance in short text modeling.
ARL-Adv is consistent with GSDMM, which has only one latent topic $z_m$ for a whole short text, as shown in Equation~\eqref{eq:z_m}. 

We denote $p(z_m|\itbold{d}_i)$ as $p_{i,z_m}$ for short.
By referring to the probability values $p_{i,z_m}$ for different $m$, it is sufficient to identify the most relevant cluster given $\itbold{d}_i$.
Besides, each probability value could be employed to rescale the representations of the corresponding clusters for $\itbold{d}_i$, yielding $\hat{\itbold{C}}_i = [\hat{\itbold{c}}_{i,1},\hat{\itbold{c}}_{i,2},\cdots,\hat{\itbold{c}}_{i,M}]$, where $\hat{\itbold{c}}_{i,m}=p_{i,z_m}\itbold{c}_m$.

\subsection{Short Text Reconstruction}\label{subsec:text-reconstruct}
Since short text clustering is indeed an unsupervised learning problem, we leverage the cluster representations to reconstruct the short text representation, which is later utilized to guide the training of the whole model.

By assuming the reconstructed representation of the $i$-th text as $\widetilde{\bm{d}}_i$, we calculate it based on a linear combination of the target-dependent cluster matrix $\hat{\itbold{C}}_i$:
\begin{align}
\widetilde{\itbold{d}}_i &= \sum_{m=1}^M \hat{\itbold{c}}_{i,m}\,.
\label{eq:di_recon}
\end{align}
Benefiting from the different contributions of clusters, we can rebuild an expressive representation of the text.
For a specific text, the derivation of cluster-level attention in Equation~\eqref{eq:z_m} encourages those clusters that are more similar to the text representation to contribute more in the linear combination in Equation~\eqref{eq:di_recon}.
When the derived reconstruction is driven to be closer to the text representation, cluster-level attention may gradually favor one single cluster as the training goes on.
The visualization in Figure~\ref{fig:visualization} can partly explain the consequence.
In the experiments, we find the average probabilities of the inferred maximum cluster are over 0.99.
These results mean each text is highly concentrated to a single cluster, rather than an arbitrary value.

\subsection{Objective Function}\label{subsec:obj-func}
Given the obtained short text representation $\itbold{d}_i$ and its reconstructed representation $\widetilde{\itbold{d}}_i$ from cluster representations, we define their relevance score to be the cosine similarity between the two representations:
\begin{align}\label{eq:relevance}
\varphi(\itbold{d}_i, \widetilde{\itbold{d}}_i) = \frac{\itbold{d}_i^\top \widetilde{\itbold{d}}_i}{\lVert\itbold{d}_i\rVert\lVert\widetilde{\itbold{d}}_i\rVert}.
\end{align}
We deem that when the relevance score gets larger, the result of the reconstruction is better.
To achieve this, we propose a hybrid objective function, consisting of two parts: a pairwise ranking part and a relevance maximization part.

As for the first part, our model seeks to minimize a margin based pairwise ranking loss~\cite{Socher14-TACL}.
Specifically, for the $i$-th short text, we sample several other texts from the corpus and regard them as a pseudo negative text set $\negativeset_i$.
And for text $D_j \in \negativeset_i$, we obtain the text similarity of the two short texts, i.e., $\varphi(\itbold{d}_i, \itbold{d}_j)$, based on the cosine similarity of their text representations calculated by Equation~\eqref{eq:relevance}, where embedding $\itbold{d}_j$ is obtained with the same procedure as shown in Equation~\eqref{eq:dt}.
Afterwards, we formulate the pairwise ranking loss for the $i$-th short text as follows:
\begin{align}
\mathcal{L}_1(i; \itbold{E}, \itbold{C}) = \frac{1}{|\negativeset_i|} \sum_{j \in\negativeset_i} \max(0, \gamma - \varphi(\itbold{d}_i, \widetilde{\itbold{d}}_i) + \varphi(\itbold{d}_i, \itbold{d}_j))\,,
\end{align}
where $|\negativeset_i|$ denotes the number of pseudo negative text (empirically setting to 16), and $\gamma$ is the margin between positive and negative ones (empirically setting to 1).
We can see minimizing the above loss would make the reconstructed $\widetilde{\itbold{d}}_i$ to have a larger relevance score with the original representation $\itbold{d}_i$ than with negative samples at the given margin.

As a supplement, we adopt the relevance maximization of $\varphi(\itbold{d}_i, \widetilde{\itbold{d}}_i)$.
To keep consistent with the optimization direction of $\mathcal{L}_1$, we define the following loss function:
\begin{align}
\mathcal{L}_2(i; \itbold{E}, \itbold{C}) = -\varphi(\itbold{d}_i, \widetilde{\itbold{d}}_i)\,,
\end{align}
which can be regarded as a pointwise loss function, aiming to predict $d_i$ as similar as possible given the combination of all cluster representations.
By fusing $\mathcal{L}_1(i)$ and $\mathcal{L}_2(i)$, it ensures that $\varphi(\itbold{d}_i, \widetilde{\itbold{d}}_i)$ not only has larger relevance than those of pseudo negative text pairs, but also a large value itself.
Taking all the short texts in corpus $\corpus$ into consideration, we define the overall objective loss function as follows:
\begin{align}
J_1(\itbold{E}, \itbold{C}) \!=\! \sum_{i=1}^{|\corpus|}\big(\mathcal{L}_1(i;\itbold{E}, \itbold{C}) \!+\! \mathcal{L}_2(i; \itbold{E}, \itbold{C})\big)\,.
\label{eq:j1}
\end{align}
From a whole perspective, ARL-Adv contains word embedding matrix $\itbold{E}$ and cluster representation matrix $\itbold{C}$ as its parameters.
Each short text is directly encoded into a low-dimensional space through its word embeddings so that no extraordinary parameters are required.

\section{Adversarial Training for Clustering}\label{sec:adversarial-train}
We propose to apply robust adversarial training to facilitate the representation learning process of ARL-Adv.
In essence, robust adversarial training provides a new objective function based on adversarial perturbations to complement the original optimization procedure.
This objective requires the model to perform well on both original samples and adversarial samples, thus benefiting the model in its training robustness, especially for short texts.

Since typical adversarial perturbations take continuous values and are added to real-valued vectors like images, they are not directly applied to discrete tokens like words.
Following the study~\cite{Miyato16-ICLR}, we add continuous adversarial perturbations to the cluster representations.
It is worth noting that we do not add adversarial perturbations to the word embedding matrix.
The reasons lie in the two aspects, i.e., efficiency and effectiveness.
First, the number of words extends far beyond that of clusters.
Thus adding perturbations to words instead of clusters entails many more parameters (perturbations) to be optimized, which might reduce the model training efficiency. 
Second, we empirically validate that applying perturbations to words is less effective than adding perturbations to cluster representations, the results of which are shown in Table~\ref{tbl:ablation}.
Formally, we define the adversarial cluster-level perturbations in ARL-Adv as $\Delta^C\in\mathbb{R}^{K \times M}$.
Consequently, we propose a new objective function formulated as follows:
\begin{align}
J_2(\itbold{E}, \itbold{C}+\Delta^C) \!=\! \sum_{i=1}^{|\corpus|}\big(\mathcal{L}_1(i;\itbold{E}, \itbold{C}+\Delta^C) \!+\! \mathcal{L}_2(i; \itbold{E}, \itbold{C}+\Delta^C)\big)\,,
\label{eq:j2}
\end{align}
where the new cluster representations $\itbold{C}+\Delta^C$, accompanied by the corresponding probability distribution $p^\Delta_{i,z_m}$ where $m\in\{1,2,\cdots,M\}$, are leveraged to constitute another reconstructed representation $\widetilde{\itbold{d}}^\Delta_i$ for short text $\corpus_i$.
The optimization of the above objective function can be viewed from two aspects.
For learning $\Delta^C$, the optimal condition is as follows:
\begin{equation}
\Delta^{C\ast} = \arg \max_{\Delta^C} J_2(\itbold{E}, \itbold{C}+\Delta^C)\,.
\label{eq:optimal_delta}
\end{equation}
On the contrary, the optimization target of $\itbold{E}$ and $\itbold{C}$ is to minimize $J_2$.
Typically, a norm-based constraint should be adopted to restrict the scale of $\Delta^C$.

By integrating the two objective functions, i.e., $J_1$ and $J_2$, the final optimization target is given as:
\begin{align}
J (\itbold{E}, \itbold{C}, \Delta^C) &= J_1(\itbold{E}, \itbold{C}) + \alpha J_2(\itbold{E}, \itbold{C}+\Delta^C)\,, \label{eq:opt-obj}  \\
\itbold{E}^\ast, \itbold{C}^\ast, \Delta^{C\ast} &= \arg \min_{\itbold{E}, \itbold{C}} \max_{\Delta^C} J(\itbold{E}, \itbold{C}, \Delta^C)\,, \label{eq:opt-param}
\end{align}
where $\alpha$ controls the relative strength of $J_2$.
The optimization of Equation~\eqref{eq:opt-obj} involves playing a minimax game.
At each step, the worst-case perturbations to cluster representations are first identified by increasing the value of $J_2$ as much as possible.
Afterwards, $\itbold{E}$ and $\itbold{C}$ are optimized to be robust to such intentional perturbations in $J_2$, while retaining satisfactory performance in $J_1$.
In such a way, for text representation $\itbold{d}_i$, its correlations with the normal reconstruction $\widetilde{\itbold{d}}_i$ and the adversarial reconstruction $\widetilde{\itbold{d}}^\Delta_i$ are simultaneously considered.
This ensures the robustness of model training, which conforms to the short-length and noisy situation of short texts.

Algorithm~\ref{algorithm1} summarizes the whole optimization process.
As usual, stochastic gradient descent algorithms such as Adam~\cite{Kingma15-ICLR} are leveraged to learn the word and clustering embeddings.
ARL-Adv calculates the gradients of $\itbold{E}$ and $\itbold{C}$ over Equation~\eqref{eq:opt-obj} through back propagation, and updates the parameters accordingly.
The learning of adversarial perturbations $\Delta^C$ follows the approximated linearizing methodology~\cite{Goodfellow15-ICLR}, leading to fast solutions with analytic forms.
With the adding of L2 norm to each  column of $\Delta^C$, i.e., $\lVert\Delta^C_m\rVert\leq\epsilon$  $(m\in \{1,2,\cdots,M\})$, we can easily derive the following updating rule for the perturbation of each cluster representation:
\begin{align}\label{eq:update-M}
\Delta^C_m = \epsilon\frac{\itbold{g}^C_m}{\lVert\itbold{g}^C_m\rVert}\,, ~~\mathrm{where}~~ \itbold{g}^C_m=\alpha\frac{\partial J_2}{\partial \itbold{C}_m}\,.
\end{align}

After finishing the training of ALR-Adv, the cluster assignments are determined based on the probability weights over the learned cluster embedding matrix $\itbold{C}$, while the adversarial perturbations $\Delta^C$ are ignored.

\begin{algorithm}
	\setlength{\belowcaptionskip}{-0.38cm}
	%\begin{small}
		\DontPrintSemicolon
		\SetKwInOut{Input}{Input}
		\SetKwInOut{Output}{Output}
		\Input{Short text corpus $\corpus$, cluster number $M$, 
			adversarial strength $\alpha$, adversarial norm $\epsilon$, and some
			other hyper-parameters such as embedding size $K$.}
			
		\Output{Word embeddings $\itbold{E}$, cluster representations $\itbold{C}$, cluster-level attention weights $p(z_m|\itbold{d}_i) (\forall m\in\{1,2,\cdots,M\}$ and $i\in\{1,2,\cdots,|\corpus|\}$).}
		
		\Begin{
			Initialize word embeddings $\itbold{E}$ and cluster representations $\itbold{C}$;\\
			\For{$iter = 1$ to $max\_iter$}{
				Sample a mini-batch $\corpus_{batch}\in \corpus$;\\
				\For{$i\in\corpus_{batch}$}{
					Constructing text pairs $(i, j)$, where $j\in\corpus_{batch}$;\\
					Learning adversarial perturbations $\Delta^C$ through Equation~\eqref{eq:update-M};\\
					Updating word embeddings $\itbold{E}$ and cluster representations $\itbold{C}$ by gradient descent of Equation~\eqref{eq:opt-param};\\
				}
			}
		}
		\caption{Robust adversarial training for ARL-Adv.}
		\label{algorithm1}
	%\end{small}
\end{algorithm}

\section{Experiments}\label{sec:exp}
In this section, we conduct experiments to answer the following pivotal questions:
\begin{itemize}
\item[\textbf{\texttt{Q1}}] Does the proposed ARL-Adv outperform standard and advanced text clustering models on short text datasets?

\item[\textbf{\texttt{Q2}}] Are the main components of the proposed model, including robust adversarial training, beneficial for the performance of short text clustering? 

\item[\textbf{\texttt{Q3}}] How good are the learned short text embeddings and cluster representations, compared to other baselines?
\end{itemize}

\subsection{Datasets}\label{subsec:datasets}   
We utilize four real-world short text datasets.
A brief introduction of each dataset and some common text preprocessing procedures are provided in the following.

\begin{itemize}[leftmargin=*]
\item[-] \textbf{TREC}\footnote{\url{https://github.com/jackyin12/MStream}\label{data12}}:
The dataset is from the Text REtrieval Conference on 2011-2015 tweet tracks.
They are organized by their corresponding queries and evaluated into several relevance levels.
We retain tweets labeled relevant or highly-relevant to their queries to ensure the quality of labels.

\item[-] \textbf{GoogleNews}\textsuperscript{\ref{data12}}:
This dataset is composed of groups of news titles and snippets clawed from Google News\footnote{\url{http://news.google.com}}.
Manual observation has confirmed its favorable grouping quality.

\item[-] \textbf{Event}:
Following the study~\cite{Ritter15-WWW}, we extract event-related tweets from an off-the-shelf tweet dataset crawled in 2016\footnote{\url{https://archive.org/}}.
Prior knowledge about the events, including the time window, relevant entities, and keywords, is fetched from Wikipedia.

\item[-] \textbf{StackOverflow}\footnote{\url{https://archive.org/download/stackexchange}}:
It is created based on the questions posted in Stack Overflow.
We require each of the selected questions to be associated with only one tag.
And the tags are regarded as the ground-truth cluster labels for the questions.
The created dataset is substantially larger than the above three datasets, hoping to provide a more convincing empirical study.
For simplicity, we sometimes use the name SO to denote this dataset.

\end{itemize}

For preprocessing, we utilize typical procedures in text processing, which consists of converting words into lowercase, removing stop words and irregular words, and applying Porter stemming.
In addition, words with frequencies below $5$ are discarded.
The detailed statistics of all datasets can be found in Table~\ref{tbl:data_stats}.
\textit{\#Cluster} and \textit{\#Text} refer to the actual number of clusters and texts for each dataset, respectively.
\textit{Vocabulary} is the number of remaining word tokens after preprocessing.
\textit{Avg. Len.} stands for the average length of texts counted in words.

\begin{table}
	\centering
	\caption{ Detailed statistics of each dataset. }
	\label{tbl:data_stats}
	\begin{tabular}{c|cccc}
	    \toprule[1.3pt]
		Dataset    &\#Cluster &\#Text &Vocabulary & Avg. Len. \\
		\hline
		TREC       &128       &11142      &3503       &7.87     \\
		GoogleNews &152       &11109      &6555       &6.23     \\
		Event      &69        &26619      &3314       &8.78     \\
		StackOverflow  &111    &137367    &10832  &5.42\\
	    \bottomrule[1.3pt]
	\end{tabular}
\end{table}

\subsection{Methods for Comparisons}\label{subsec: method-comparison}

\subsubsection{Baselines}
The selected representative baselines are categorized into conventional text clustering methods (\romannumeral1), deep learning based general clustering models (\romannumeral2), and tailored methods for short text clustering (\romannumeral3).
The details of the baselines are given below for clarity.

\begin{itemize}[leftmargin=*]

\item[-] \textbf{K-means}:
It is a classical and simple clustering approach, relying on hand-crafted features for text clustering.
Two feature types are adopted in our experiments, i.e., TF-IDF and low-dimensional representations obtained by PCA.
We denote K-means running on these two different features as K-means(TF-IDF) and K-means(PCA), respectively.
For K-means(PCA), we have tried the number of primary components in $\{100, 200, 300\}$.
It turns out $200$ is the more suitable choice, so we report results based on this setting.
The Euclidean distance is utilized as the internal similarity metric.
For each run, $10$ random centroids are initialized.

\item[-] \textbf{HieClu}:
Similar as K-means, Hierarchical Clustering (HieClu) is another simple baseline for clustering.
The results of taking textual features from PCA are reported due to its better performance than TF-IDF.

\item[-] \textbf{LDA}~\cite{Blei03-JMLR}:
LDA is a classical and standard generative statistical model which learns a topic (cluster) distribution for each document.
We assign a text with the topic that has the largest probability value inferred by LDA.
Following~\cite{Rosen04-UAI}, we set $\alpha=K/50$ and $\beta=0.1$, where $K$ is the number of topics.
Other parameter settings are tried as well, but we observe no significant improvement in performance.

\item[-] \textbf{BTM}~\cite{Yan13-WWW}\footnote{\url{https://github.com/xiaohuiyan/BTM}}:
BTM regards bi-grams as the representations of short texts and generates them conditioned on different topics.
Similar to LDA, we tune $\alpha=K/50$ and set $\beta=0.05$ based on performance.
The number of training iterations is set to $500$ for BTM to reach convergence.

\item[-] \textbf{GSDMM}~\cite{Yin14-KDD}\footnote{\url{https://github.com/jackyin12/GSDMM}}:
GSDMM is tailored for short text clustering and assumes that each text is generated by one topic, which is fundamentally different from LDA that generates one word from one topic.
We set the parameters $\alpha=K/50$ and $\beta=0.05$, and train the model for $300$ iterations.

\item[-] \textbf{DEC}~\cite{Xie16-ICML}\footnote{\url{https://github.com/piiswrong/dec}}:
It is a deep embedded clustering model that leverages autoencoder, with TF-IDF features as input to map documents into low-dimensional embeddings.
Then the mapping function and cluster representations are refined based on the idea of self-training.
Similar to VaDE, DEC has specified transforming and parameter settings for text data.
We perform pre-training for $500$ iterations, and training for $10000$ iterations to ensure its convergence.
SGD is the optimizer during both pre-training and training phases, with momentum set to $0.9$.
The batch size is set to $128$.
Empirical results show that setting the dimension of the hidden layer in DEC to $80$ can help it get consistent and robust performance.

\item[-] \textbf{STCC}~\cite{Xu15-NAACL,Xu17-NN}\footnote{\url{https://github.com/jacoxu/STC2}}:
This model mainly consists of three separate steps.
It first trains a convolutional neural network with the help of autoencoders.
Afterwards, the well-trained model is employed to get text embeddings, which are fed into K-means for final clustering.
We set the kernel width to $3$ for convolutional layers, with $k=5$ in its $k$-$max$ pooling.
The number of feature maps on the first and second convolutional layer is 32 and 16, respectively.
Adam is used for batch based optimization, with the learning rate of $5e$-$3$ and batch size of $100$.

\item[-] \textbf{VaDE}~\cite{Jiang17-IJCAI}\footnote{\url{https://github.com/slim1017/VaDE}}:
VaDE extends canonical variational auto-encoding approaches to support clustering tasks by utilizing the idea of Gaussian mixture model.
We partially follow the parameter settings for text data in the implementation of VaDE and keep the procedure of TF-IDF based feature transformation.
Setting pre-training to $500$ iterations and training to $10000$ iterations results in the convergence of VaDE on an empirical basis.
Both phases use Adam as the optimizer, with the learning rate of $1e$-$3$.
The default configuration of Adam is adopted, i.e., $\beta_1=0.9$ and $\beta_2=0.999$.
We adjust the size of the hidden representations to $40$, and batch size to $128$ for more general and stable outcomes.

\item[-] \textbf{ClusterGAN}~\cite{MukherjeeALK19}\footnote{\url{https://github.com/sudiptodip15/ClusterGAN}}:
ClusterGAN is recently proposed to leverage GANs to encode continuous data to discrete cluster labels.
Considering its requirement of continuous data representations, we have tried TF-IDF and word2vec (average word embedding) to denote each short text, and reported the better performance.

\item[-] \textbf{STC}~\cite{Hadifar19-ACL}\footnote{\url{https://github.com/hadifar/stc_clustering}}:
STC adopts self-training inspired by DEC and uses Smoothed Inverse Frequency (SIF) to compute a weighted average of pre-trained word embeddings in a stage independent of optimizing its clustering model.
The word embeddings are initialized by word2vec (the same as ARL-Adv), and their size is tuned to be 32.
The dimension of its autoencoder is set to 80.
Other parameters are kept the same as their original settings.
\end{itemize}

All the above models are tuned for the experimental datasets.
To ensure statistical significance, all results in the experiments are averaged over ten runs.
We also tried GaussianLDA\footnote{\url{https://github.com/rajarshd/Gaussian_LDA}} and GANMM\footnote{\url{https://github.com/eyounx/GANMM}} introduced in Section~\ref{subsec:text-clu}.
However, the results show that GaussianLDA could not obtain competitive clustering performance than other baselines.
And training GANMM incurs a heavy computational burden, especially when the cluster number is not small.
Therefore we omit their details in this paper.

\begin{table}[!t]
	\renewcommand{\arraystretch}{1.2}
	\centering
	\caption{Default settings for ARL-Adv. }
	\label{tbl:hyper-setting}
	\begin{tabular}{c|c|c|c}
	    \toprule[1.3pt]
		$K$ & $|\corpus_{batch}|$ & $\alpha$ & $\epsilon$ \\
		\hline 
        300 & 64 & 1.0 & 50.0\\
	    \bottomrule[1.3pt]
	\end{tabular}
\end{table}

\begin{table*}
	\centering
	\caption{ Performance comparison over all methods. }
	\label{tbl:performance}
	\begin{tabular}{c|c|ccc|ccc|ccc|ccc}
	    \toprule[1.3pt]
		\multirow{2}{*}{Type}
		&Dataset
		&\multicolumn{3}{c|}{TREC}  &\multicolumn{3}{c|}{GoogleNews}  &\multicolumn{3}{c|}{Event}
		&\multicolumn{3}{c}{StackOverflow} \\
		&Metrics 
		&NMI     &ARI     &ACC     &NMI     &ARI     &ACC     &NMI     &ARI     &ACC     &NMI     &ARI &ACC  \\
		\hline
		\multirow{3}{*}{\romannumeral1}
        &LDA
		&0.7514  &0.5897  &0.6120  &0.7197  &0.5074  &0.5878  &0.6348  &0.4557  &0.4984
		&0.2121  &0.1044  &0.2049 \\
		&K-means(TF-IDF)
		&0.8376  &0.3682  &0.6463  &0.7941  &0.2316  &0.5897  &0.6790  &0.3290  &0.4473  &0.3110  &0.0682  &0.2459\\
		&K-means(PCA)
		&0.8289  &0.3460  &0.6267  &0.7724  &0.2011  &0.5547  &0.6680  &0.3150  &0.4332
		&0.2885  &0.0848  &0.2319 \\
        &HieClu
		&0.8567  &0.4975  &0.7118  &0.7816  &0.2459  &0.5897  &0.5901  &0.2346  &0.4013
		&0.2756  &0.0336  &0.2229\\		
		\hline
		\multirow{2}{*}{\romannumeral2}
		&ClusterGAN
		&0.6445&0.3602&0.4953
		&0.7206&0.4058&0.5396
		&0.4061&0.1211&0.2247
		&0.2217&0.0844&0.2038\\
		&DEC
		&0.8664  &0.5839  &0.6993  &0.8505  &0.5088  &0.6705  &0.7423  &0.4264  &0.5320
		&0.3799  &0.1516  &0.3186 \\
		&VaDE
		&0.8712  &0.6632  &0.7170  &0.8389  &0.5318  &0.6287  &0.6620  &0.2927  &0.4078  
		&0.3214  &0.1277  &0.2670 \\
		\hline
		\multirow{5}{*}{\romannumeral3}
		&STCC
		&0.8172  &0.5758  &0.6455  &0.7947  &0.5262  &0.6122  &0.6567  &0.3797  &0.3959  
		&0.3608  &0.1433  &0.2522 \\
		&BTM
		&0.8759  &0.6920  &0.7182  &0.8656  &0.6310  &0.7023  &0.6781  &0.3834  &0.4704  
		&0.3508  &0.1537  &0.2565 \\
		&GSDMM
		&0.8746  &0.7453  &0.7512  &0.8700  &0.6782  &0.7278  &0.7572  &0.4937  &0.5039  
		&0.3569  &0.1418  &0.2598 \\
		&STC
		&0.8906  &0.7372  &0.7589  &0.8667  &0.6471  &0.7169  &0.6431  &0.3170  &0.3982 
		&0.3903  &0.1598  &0.2750 \\
		\cline{2-14} 
		&\textbf{ARL}
		&0.9261  &0.8428  &0.8347  &0.8995  &0.7897  &0.8057  &0.8243  &0.6078  &0.6156  
		&0.4796  &0.2731  &0.3781 \\
		&\textbf{ARL-Adv}
		&\textbf{0.9305}  &\textbf{0.8486}  &\textbf{0.8402}  
		&\textbf{0.9054}  &\textbf{0.8214}  &\textbf{0.8303}  
		&\textbf{0.8450}  &\textbf{0.6728}  &\textbf{0.6607} 
		&\textbf{0.4871}  &\textbf{0.3006}  &\textbf{0.3943}
		\\
	    \bottomrule[1.3pt]
	\end{tabular}
\end{table*}

\subsubsection{Variants of the proposed model}\label{subsubsec:arl_variants}
To verify the effectiveness of the main components in the proposed model, we consider some of its variants, which are introduced below.

\begin{itemize}[leftmargin=*]
\item[-] \textbf{ARL-Adv}:
This is the full version of our proposed model, which adopts the optimization target as Equation~\eqref{eq:opt-obj} and enables adversarial training.
\item[-] \textbf{ARL}:
Only $J_1$ is employed to train the model.
By comparing ARL with the full model ARL-Adv, we can verify the effectiveness of robust adversarial training for the unsupervised clustering task.
\item[-] \textbf{ARL-Adv(no train w)}:
This method optimizes parameters other than the word embedding matrix.
Comparison against it can show the necessity of learning word-level semantic representations.
\item[-] \textbf{ARL-Adv(no train c)}:
The cluster representation matrix is not learned in this model, hoping to partially verify the benefit of combining representation learning and clustering.
\item[-] \textbf{ARL-Adv w/o $\mathcal{L}_1$}:
The pairwise ranking loss $\mathcal{L}_1$ is removed from ARL-Adv.
That is, the short texts do not interact with their negative samples.
\item[-] \textbf{ARL-Adv w/o $\mathcal{L}_2$}:
It means ARL-Adv does not consider the supplementary pointwise loss $\mathcal{L}_2$.
Both ARL-Adv w/o $\mathcal{L}_1$ and ARL-Adv w/o $\mathcal{L}_2$ still perform adversarial training.
\item[-] \textbf{ARL-Random}:
This variant takes random noise as perturbations, with the same scale level as intentional adversarial perturbations.
\item[-] \textbf{ARL-Adv(word)}:
It shares a very similar spirit with ARL-Adv, except that adversarial perturbations are added to word embeddings instead of cluster representations.
\end{itemize}

For ARL-Adv and its variants, word embeddings are pre-trained on each dataset with word2vec\footnote{\url{https://radimrehurek.com/gensim/models/word2vec.html}} for $200$ iterations.
Moreover, we initialize centroids of the clusters by performing K-means on short text embeddings, the same as the study~\cite{Xie16-ICML}.
Without loss of generality, the default hyper-parameters are set to the ones shown in Table~\ref{tbl:hyper-setting}.
$K$ is the dimension of embeddings, and $|\corpus_{batch}|$ is the size of a batch.
Unless otherwise stated, the results will be reported under this setting.

\subsection{Evaluation Metrics}\label{sec:metrics}
We adopt three commonly used metrics for text clustering performance evaluation, i.e., normalized mutual information (NMI)~\cite{Strehl02-JMLR}, adjusted rand index (ARI)~\cite{Hubert85-JC}, and clustering accuracy (ACC)~\cite{Xie16-ICML}.
Suppose $\nu_i$ denotes the number of short texts in the $i$-th true topic, and $\widetilde{\nu}_j$ represents the number of short texts in the $j$-th inferred cluster.
We use $\bar{\nu}_{ij}$ to denote the number of short texts simultaneously appearing in two clusters.
In addition, $\mathcal{M}$ denotes the set of all possible one-to-one mappings between the generated clusters and real topics.
For an efficient search of the best mapping, Hungarian algorithm~\cite{Kuhn55-NRL} can be adopted.
Formally, NMI, ARI, and ACC can be formulated as follows:
\begin{equation}
	\mathrm{NMI} = \frac
	{\sum_{ij} \bar{\nu}_{ij} \log\frac{|\corpus| \cdot \bar{\nu}_{ij}}{\nu_i \widetilde{\nu}_j}}
	{\sqrt{
    (\sum_i \nu_i \log\frac{\nu_i}{|\corpus|} )
    (\sum_j \widetilde{\nu}_j \log\frac{\widetilde{\nu}_j}{|\corpus|})}}\,,
\end{equation}
\begin{equation}
	\mathrm{ARI} = \frac
	{\sum_{ij}\!\binom{\bar{\nu}_{ij}}{2} \!\!-\!\! \left[ \sum_i\!\binom{\nu_i}{2} \sum_j\!\binom{\widetilde{\nu}_j}{2} \right] \!/\! \binom{|\corpus|}{2}}
	{\frac{1}{2} \!\! \left[\sum_i\!\binom{\nu_i}{2} \!+\! \sum_j\!\binom{\widetilde{\nu}_j}{2} \right] \!\!-\!\! \left[ \sum_i\!\binom{\nu_i}{2} \sum_j\!\binom{\widetilde{\nu}_j}{2} \right] \!/\! \binom{|\corpus|}{2}}\,,
\end{equation}
\begin{equation}
    \mathrm{ACC} = \max_{m\in\mathcal{M}} \frac{\sum_{i=1}^{|\corpus|} \mathbf{1}\{l_i==m(c_i)\}}{|\corpus|}\,.
\end{equation}

Higher values evaluated by NMI, ARI, and ACC indicate better clustering quality.
They are all equal to 1 when a perfect match is achieved in cluster assignments on a whole corpus.
Both NMI and ARI penalize unnecessary splits of texts from the same true cluster to several inferred clusters.
ARI further penalizes undesirable merging of texts from different true clusters into the same inferred cluster, making it a more rigorous metric.

\subsection{Software Configuration}\label{exp-sf}
We conduct all the experiments on a server with 32 CPUs and 4 GPUs (GTX1080Ti).
For all baselines, the software environments are configured according to their specific requirements.
Extensively used in the literature, Scikit-learn is adopted for the implementations of K-means, HieClu, and LDA, the visualization tool t-SNE, and the metrics NMI, ARI, and ACC.
Our models ARL and ARL-Adv are built based on Python and Tensorflow, with CUDA and cuDNN to enable the usage of GPUs.
We release the Event dataset\footnote{\url{https://github.com/CodeUpload-E/ARL-Adv}} described in Section~\ref{subsec:datasets}. 
With the publication of this paper, the source code will be prepared for relevant studies.

\begin{table*}
	\centering
	\caption{Ablation study of the proposed model.}
	\label{tbl:ablation}
	\begin{tabular}{c|ccc|ccc|ccc|ccc}
	    \toprule[1.3pt]
		Dataset
		&\multicolumn{3}{c|}{TREC}  &\multicolumn{3}{c|}{GoogleNews}  &\multicolumn{3}{c}{Event}  
		&\multicolumn{3}{c}{StackOverflow}  \\
		Metrics
		&NMI     &ARI     &ACC     &NMI     &ARI     &ACC     &NMI     &ARI     &ACC     &NMI     &ARI &ACC  \\
		\hline
		ARL-Adv(no train w)
		&0.9001  &0.7257  &0.7631  &0.8699  &0.6483  &0.6991  &0.7039  &0.3783  &0.4450  
		&0.3871     &0.1819 &0.2917  \\
		ARL-Adv(no train c)
		&0.9029  &0.7755  &0.7954  &0.8783  &0.7335  &0.7688  &0.7740  &0.6230  &0.6072  
		&0.4787     &0.2870 &0.3859  \\
		\hline
		ARL-Adv w/o $\mathcal{L}_1$
        &0.9279  &0.8408  &0.8368  &0.9005  &0.7881  &0.8044  &0.6826  &0.3915  &0.4631 
        &0.3483     &0.0942 &0.2294  \\
		ARL-Adv w/o $\mathcal{L}_2$
        &0.9286  &0.8441  &0.8380  &0.9033  &0.8120  &0.8240  &0.8176  &0.6244  &0.6297 
        &\textbf{0.4946}     &\textbf{0.3071} &\textbf{0.4021}  \\
        \hline
		ARL
		&0.9261  &0.8428  &0.8347  &0.8995  &0.7897  &0.8057  &0.8243  &0.6078  &0.6156 
		&0.4796     &0.2731 &0.3781  \\
		ARL-Random
		&0.9250  &0.8395  &0.8330  &0.8980  &0.7813  &0.7998  &0.8174  &0.6005  &0.6064  
		&0.4761     &0.2564 &0.3678  \\
		ARL-Adv(word)
		&0.9271  &0.8458  &0.8358  &0.9014  &0.7925  &0.8074  &0.8294  &0.6387  &0.6307  
		&0.4793     &0.2604 &0.3684  \\
		ARL-Adv
		&\textbf{0.9305}  &\textbf{0.8486}  &\textbf{0.8402}  
		&\textbf{0.9054}  &\textbf{0.8214}  &\textbf{0.8303}  
		&\textbf{0.8450}  &\textbf{0.6728}  &\textbf{0.6607}  &0.4871     &0.3006 &0.3943  \\
	    \bottomrule[1.3pt]
	\end{tabular}
\end{table*}

\subsection{Model Comparison (\textbf{\texttt{Q1}})}\noindent
We report the evaluation results in Table~\ref{tbl:performance} for ARL, ARL-Adv, and all the baselines.
The cluster numbers here are set to be the same as those in Table~\ref{tbl:data_stats}.
From a whole perspective, all the methods do not perform the same w.r.t. the three metrics.
For example, K-means(TF-IDF) gains better results than LDA w.r.t. NMI and ACC, but exhibits worse performance w.r.t. ARI.
This phenomenon shows the necessity of adopting all the three metrics.
Since they focus on different properties of the clustering results, using all of them can provide more comprehensive comparisons from distinct perspectives.

As expected, the models belonging to the conventional category (\romannumeral1) do not show competitive performance than other types of baselines in most cases, since they are not tailored for short texts and do not fully leverage the power of representation learning.
Specifically, LDA works poorly here because assigning different clusters to words in the same short text can aggravate sparsity, as the corresponding results demonstrate.
Although PCA converts the sparse representation of TF-IDF to continuous low-dimensional space, K-means(PCA) shows no improvements over K-means(TF-IDF).
This result indicates that introducing representation that is independent of the model training process may not easily work.
Besides, although the classical and simple model HieClu adopts another mechanism (i.e., agglomerative clustering), its performance is still at the same level as K-means.

Both DEC and VaDE are deep learning based approaches (\romannumeral2), which can learn low-dimensional representations of targets using their original representations, such as the pixel values of images and TF-IDF based representations of texts.
Therefore, though proposed for general clustering tasks, they may not be limited to texts.
As shown in Table~\ref{tbl:performance}, they yield better results in most cases than the category of conventional methods.
By further investigating the results of ClusterGAN, we find it does not exhibit the satisfied clustering performance.
This might be attributed to the reason that the model is incapable of learning representations for discrete text tokens due to its GAN-based design.

Among the baselines originally proposed for short text clustering (\romannumeral3), STCC performs not very well.
This comparison shows that separating the representation learning and clustering into different stages tends to be sub-optimal.
The reason might be that the representation learning process lacks proper guidance from the feedback of clustering if they are not learned together.
Compared to STCC, GSDMM and BTM perform much better in most cases.
In particular, GSDMM outperforms the general deep learning based clustering models in most cases.
We attribute this phenomenon to the proper assumption in GSDMM that short texts are generated from only one topic, along with its proper approximation in the posterior distribution.
STC exhibits good performance of TREC and StackOverflow (still worse than ARL-Adv) but not so good results on the other two datasets.
It makes sense since STC cannot optimize word embeddings when training its clustering method.

To sum up, our full model ARL-Adv is superior since its improvements over the baselines are statistically significant on all the datasets, verified by t-test.
This might be attributed to the following reasons:
(1) ARL-Adv utilizes word embeddings to obtain short text embeddings and learns word embeddings with the optimization of the whole model rather than learning it separately.
(2) ARL-Adv combines the representation learning and short text clustering in an end-to-end learning fashion.
(3) Robust adversarial training adopted in ARL-Adv can effectively improve clustering performance.
The last part of Table~\ref{tbl:performance} shows that ARL-Adv improves ARL consistently in the four datasets, especially for GoogleNews, Event, and StackOverflow, verifying the third reason.
In the next section, we empirically demonstrate the benefit of learning word embeddings and cluster representations by the ablation study on the model.

\begin{figure*}[!t]
    \centering
	\subfigure[(TREC) TF-IDF]{
		\begin{minipage}[t]{0.16\linewidth} 
		\centering \includegraphics[scale=0.08]{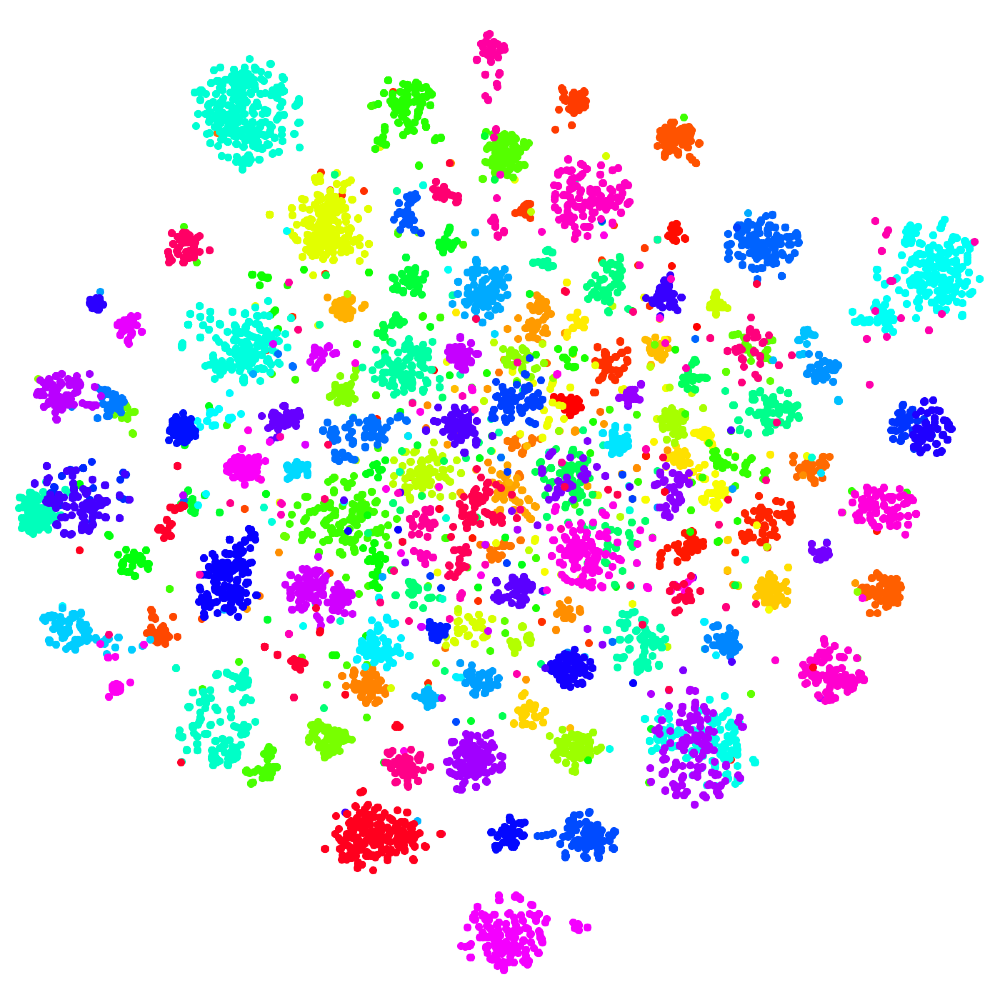} \end{minipage}
	}
	\subfigure[(TREC) VaDE]{
		\begin{minipage}[t]{0.16\linewidth} 
		\centering \includegraphics[scale=0.08]{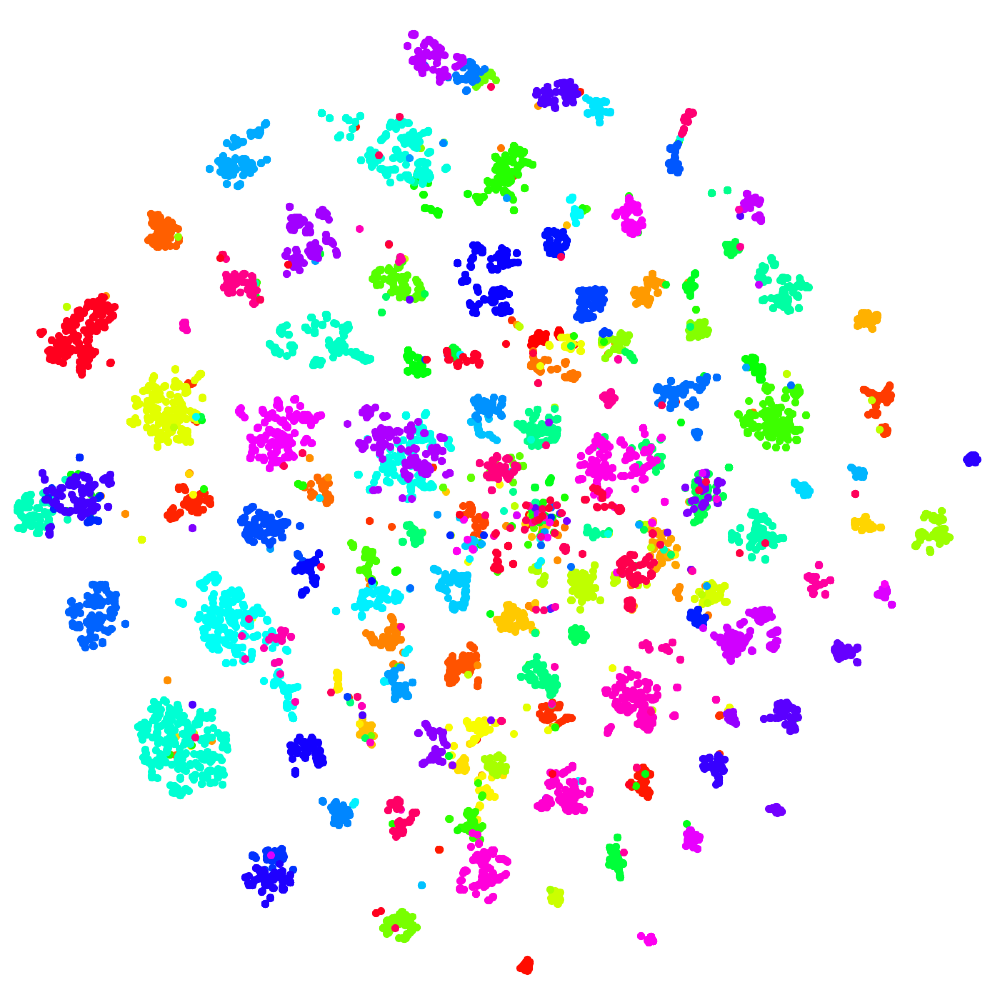}
		\end{minipage}
	}
	\subfigure[(TREC) DEC]{
    	\begin{minipage}[t]{0.16\linewidth}
		\centering\includegraphics[scale=0.08]{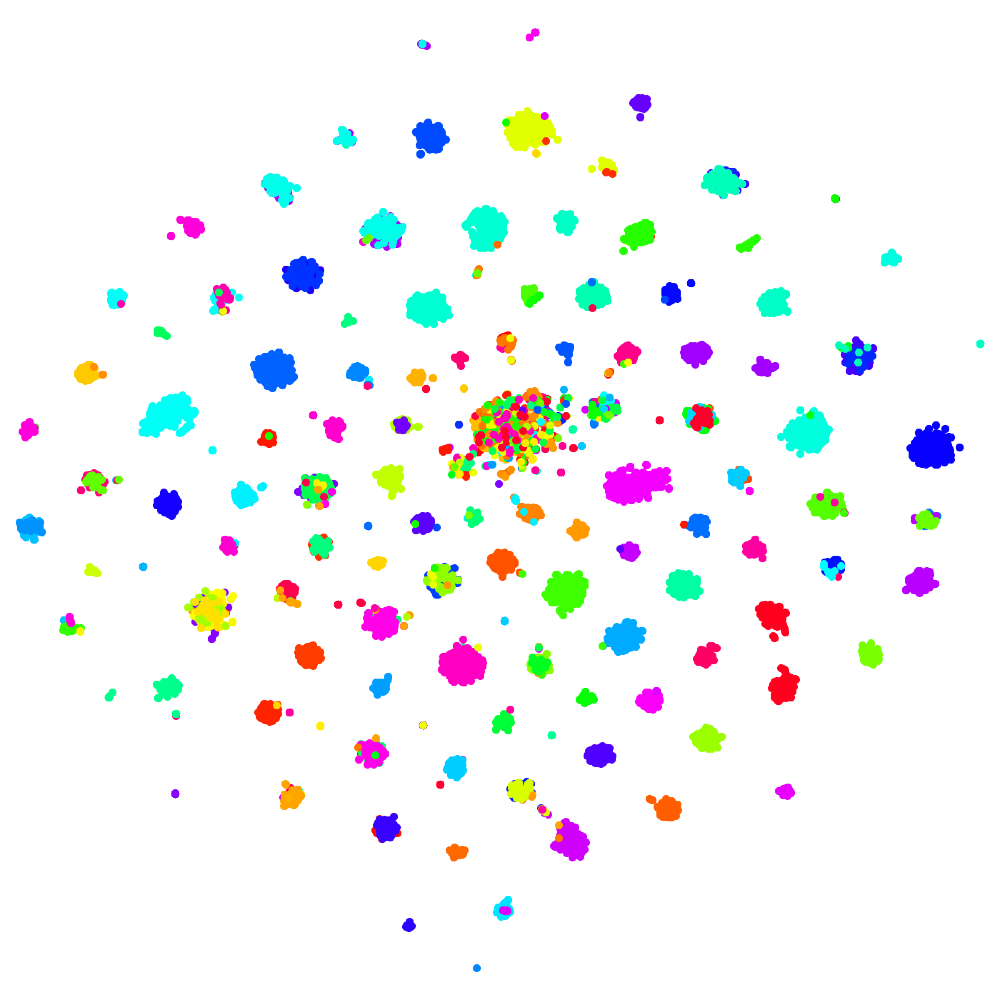}
        \end{minipage}
	}
	\subfigure[(TREC) ARL]{
    	\begin{minipage}[t]{0.16\linewidth}
		\centering \includegraphics[scale=0.08]{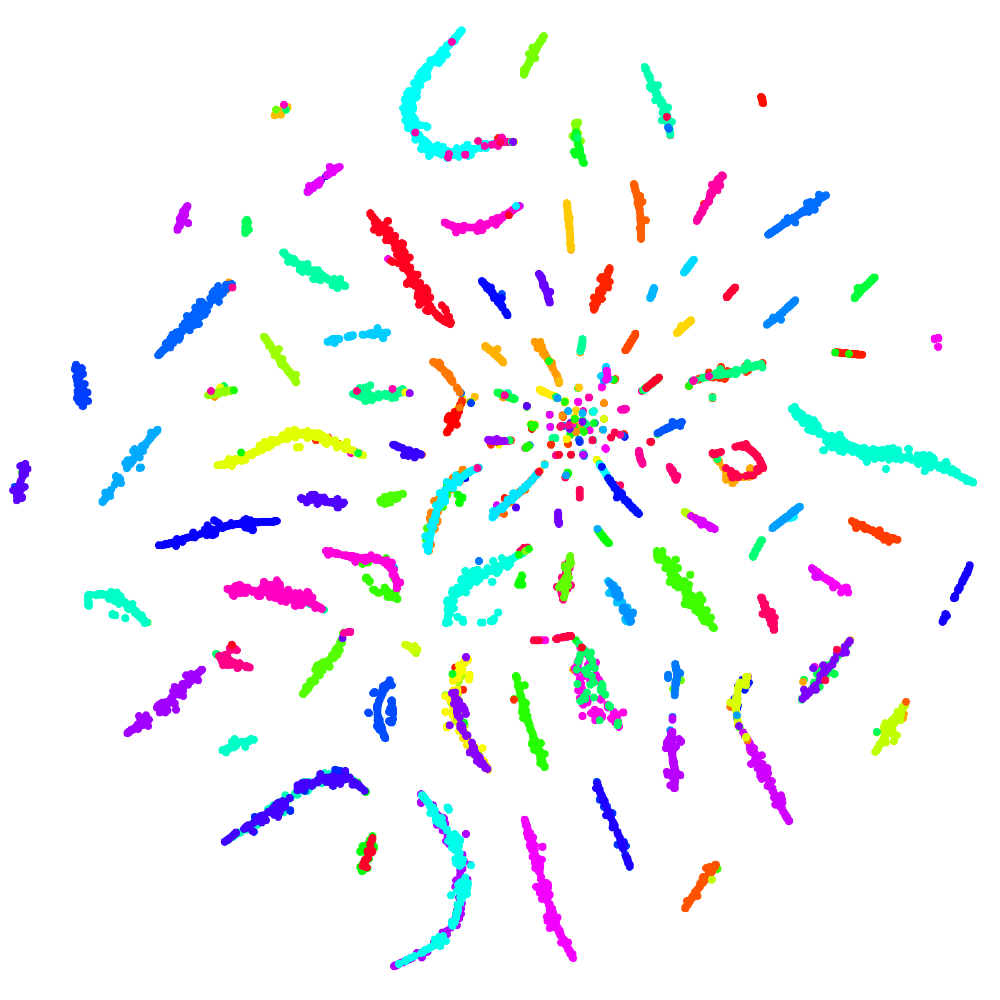}
        \end{minipage}
	}	
	\subfigure[(TREC) ARL-Adv]{
    	\begin{minipage}[t]{0.16\linewidth} 
		\centering \includegraphics[scale=0.08]{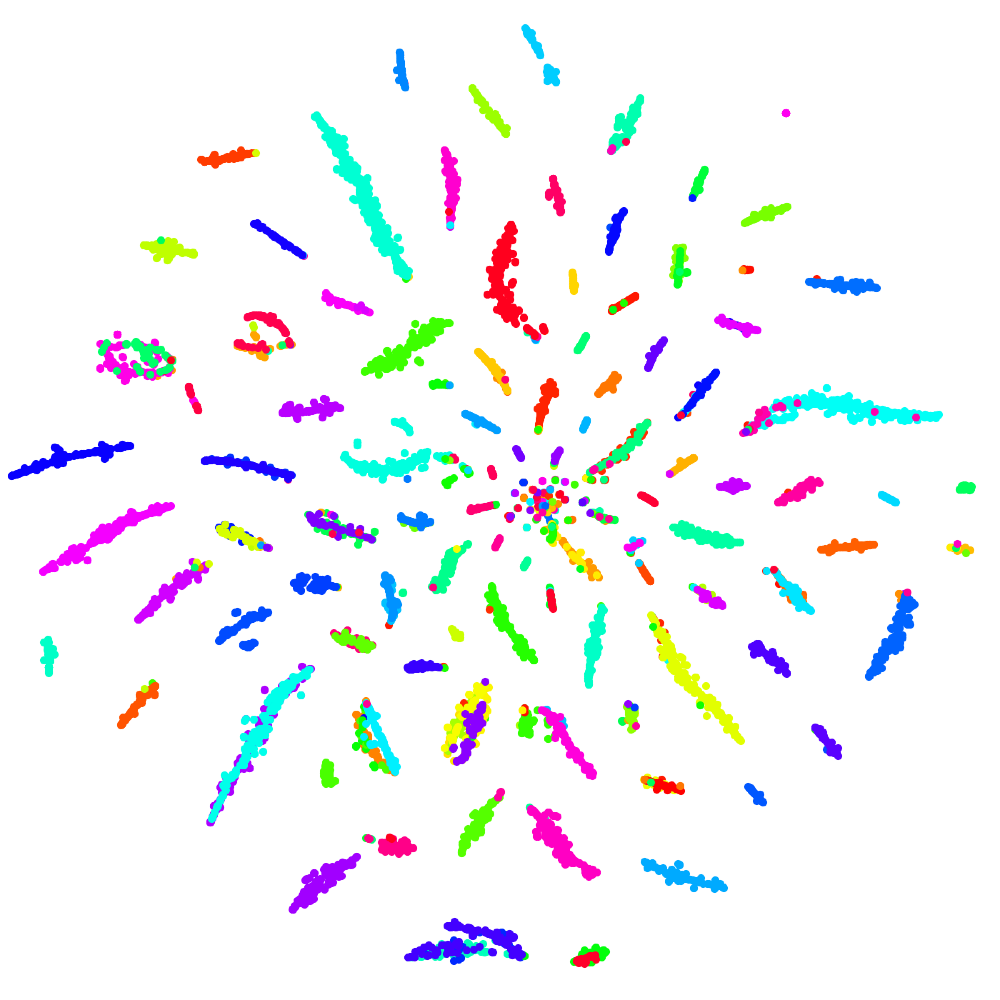}
        \end{minipage}
	}
	\subfigure[(GoogleNews) TF-IDF]{
    	\begin{minipage}[t]{0.16\linewidth}
		\centering \includegraphics[scale=0.08]{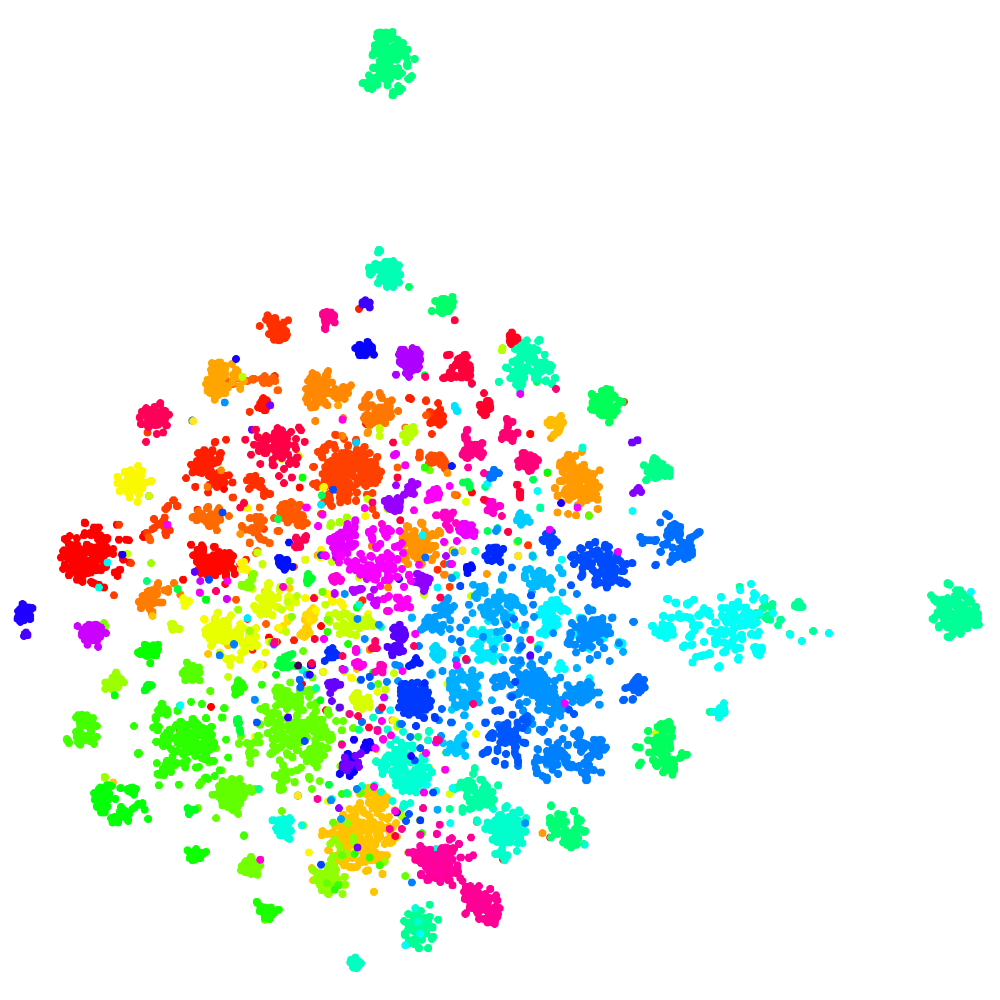}
        \end{minipage}
	}
	\subfigure[(GoogleNews) VaDE]{
    	\begin{minipage}[t]{0.16\linewidth} 
		\centering \includegraphics[scale=0.08]{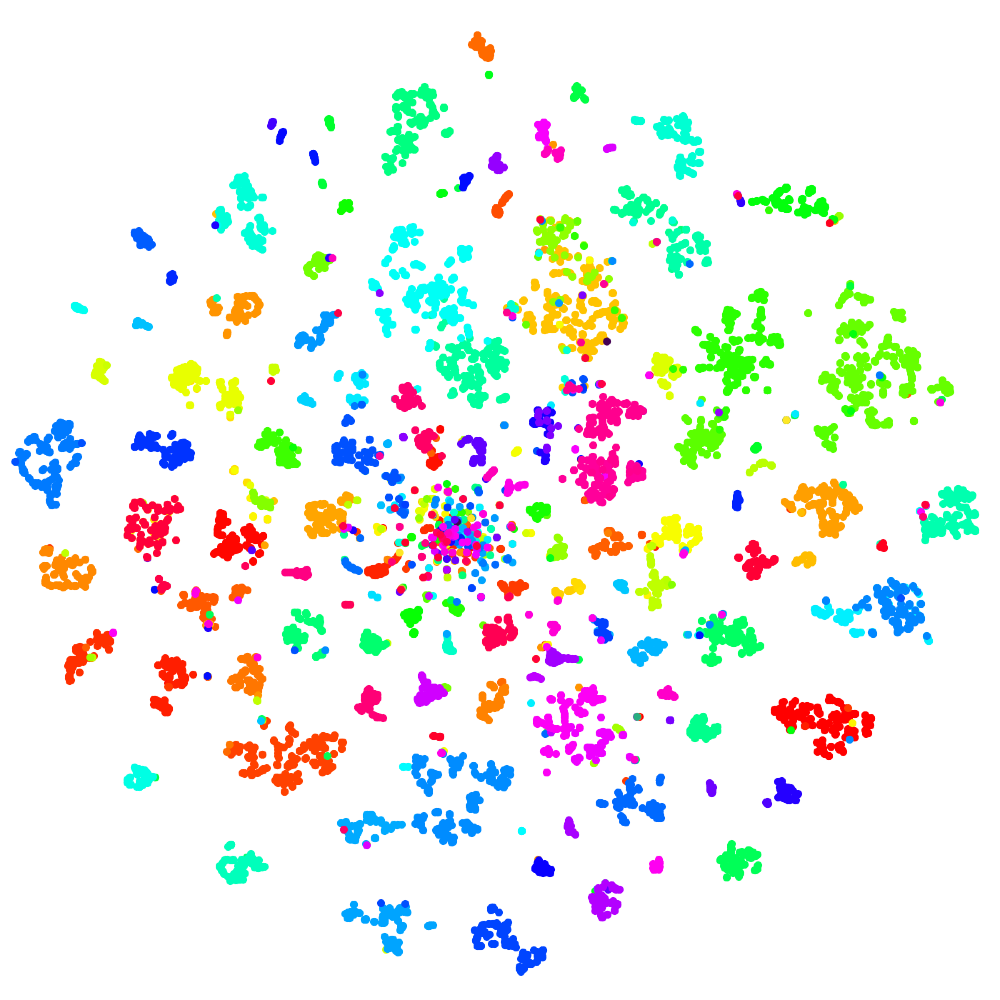}
        \end{minipage}
	}
	\subfigure[(GoogleNews) DEC]{
    	\begin{minipage}[t]{0.16\linewidth} 
		\centering \includegraphics[scale=0.08]{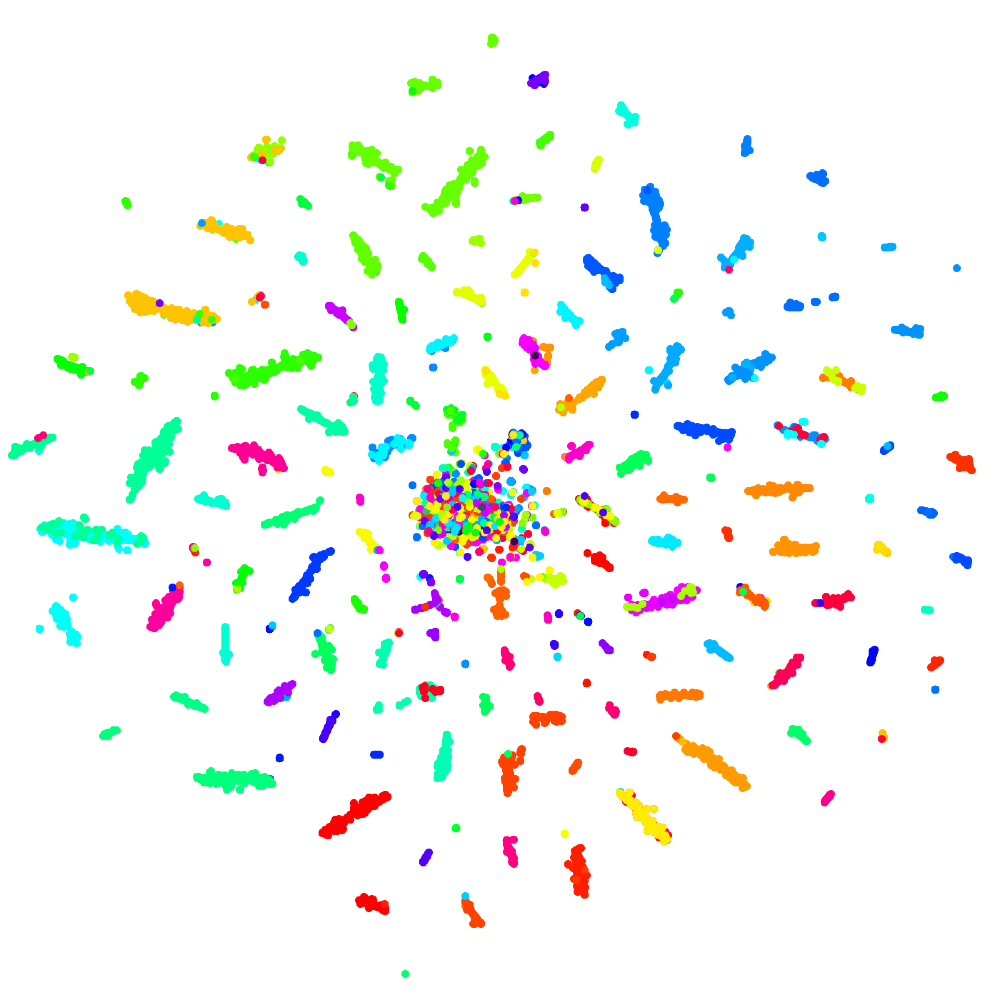}
        \end{minipage}
	}
	\subfigure[(GoogleNews) ARL]{
    	\begin{minipage}[t]{0.16\linewidth} 
		\centering \includegraphics[scale=0.08]{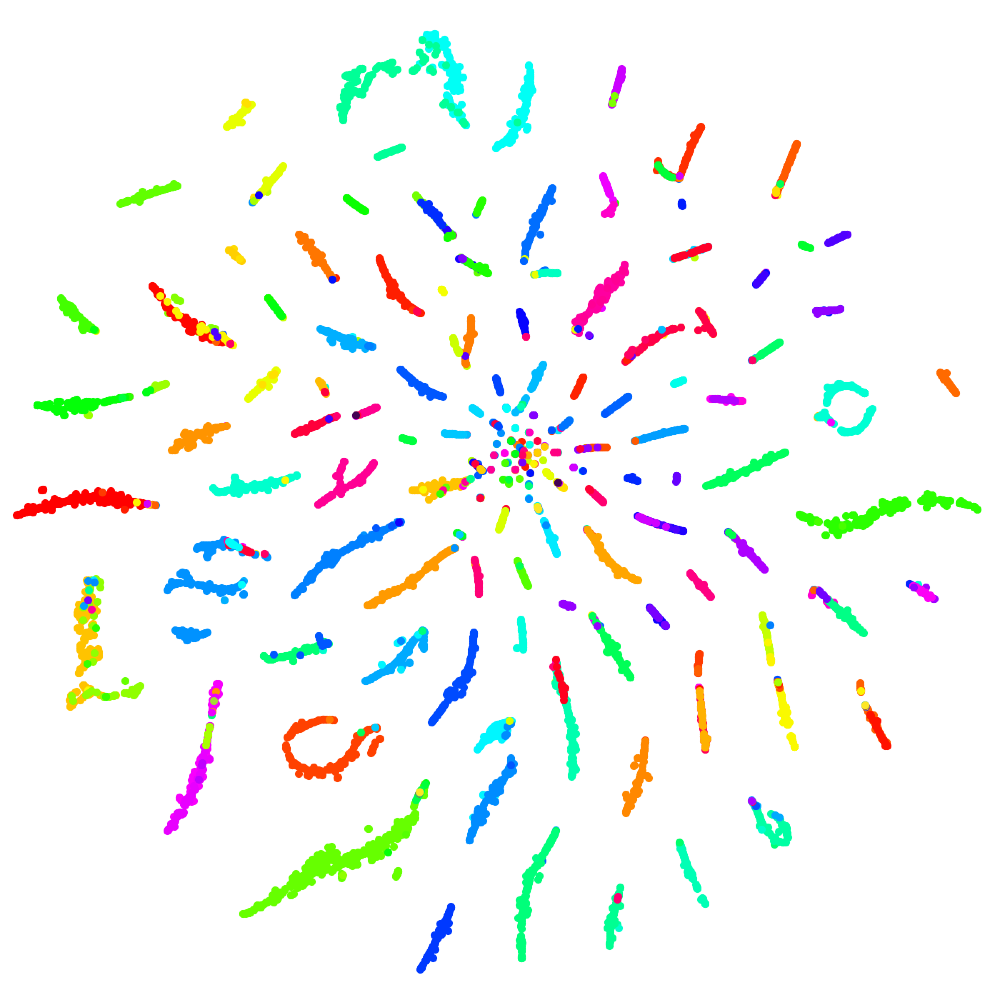}
        \end{minipage}
	}
	\subfigure[(GoogleNews) ARL-Adv]{
    	\begin{minipage}[t]{0.17\linewidth} 
		\centering \includegraphics[scale=0.08]{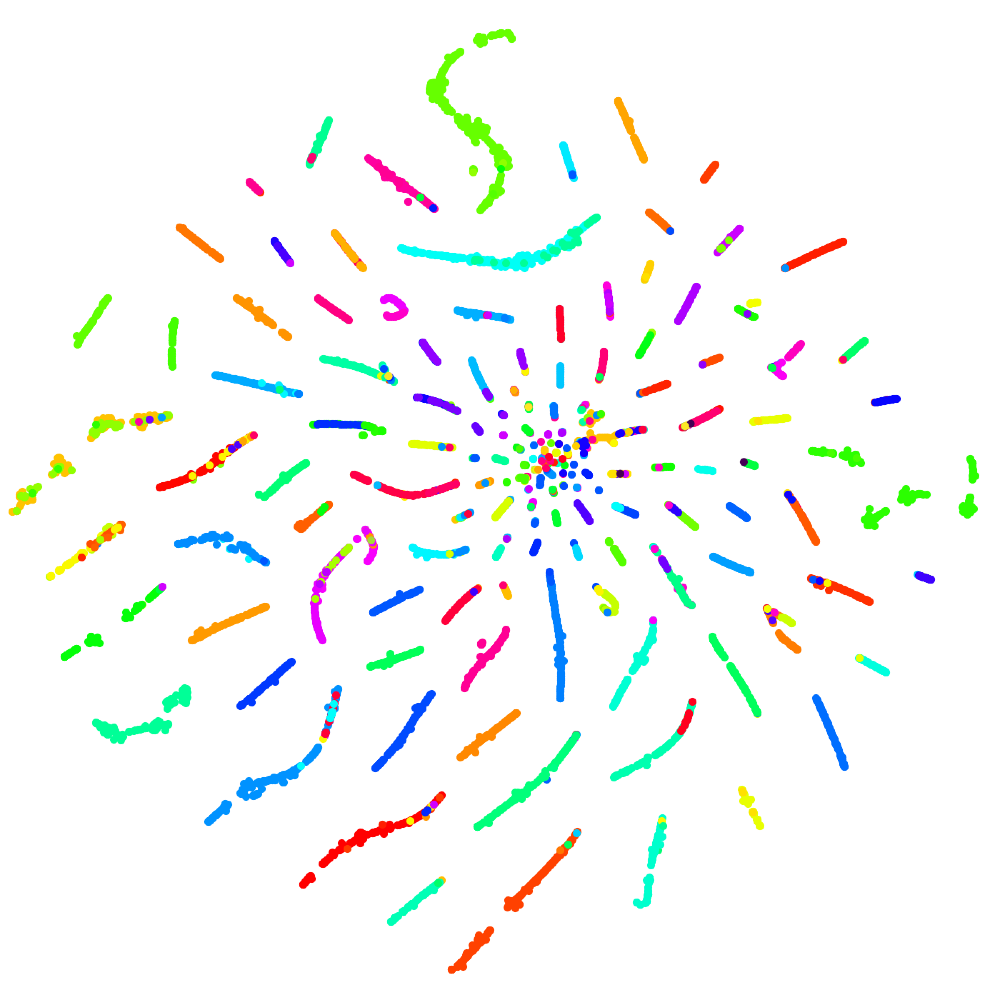}
        \end{minipage}
	}
	\subfigure[(Event) TF-IDF]{
    	\begin{minipage}[t]{0.16\linewidth} 
		\centering \includegraphics[scale=0.08]{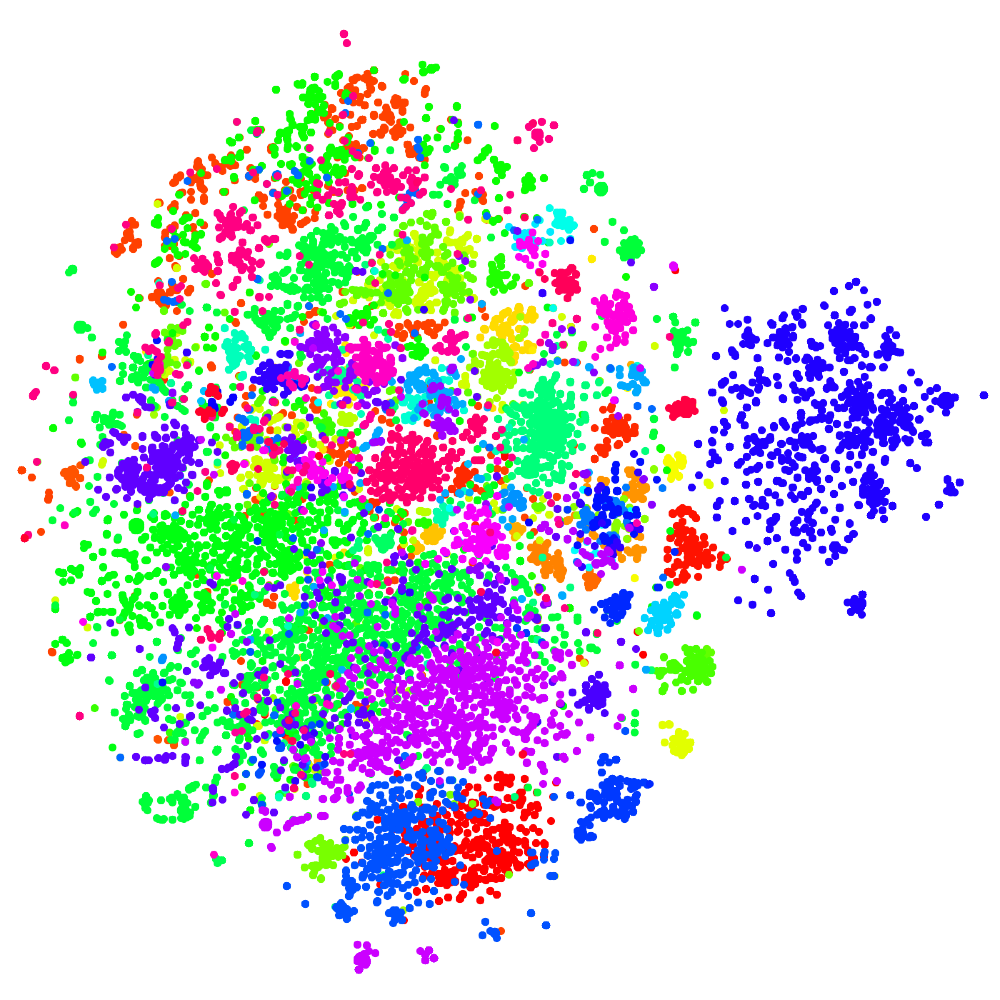}
        \end{minipage}
	}
	\subfigure[(Event) VaDE]{
    	\begin{minipage}[t]{0.16\linewidth} 
		\centering \includegraphics[scale=0.08]{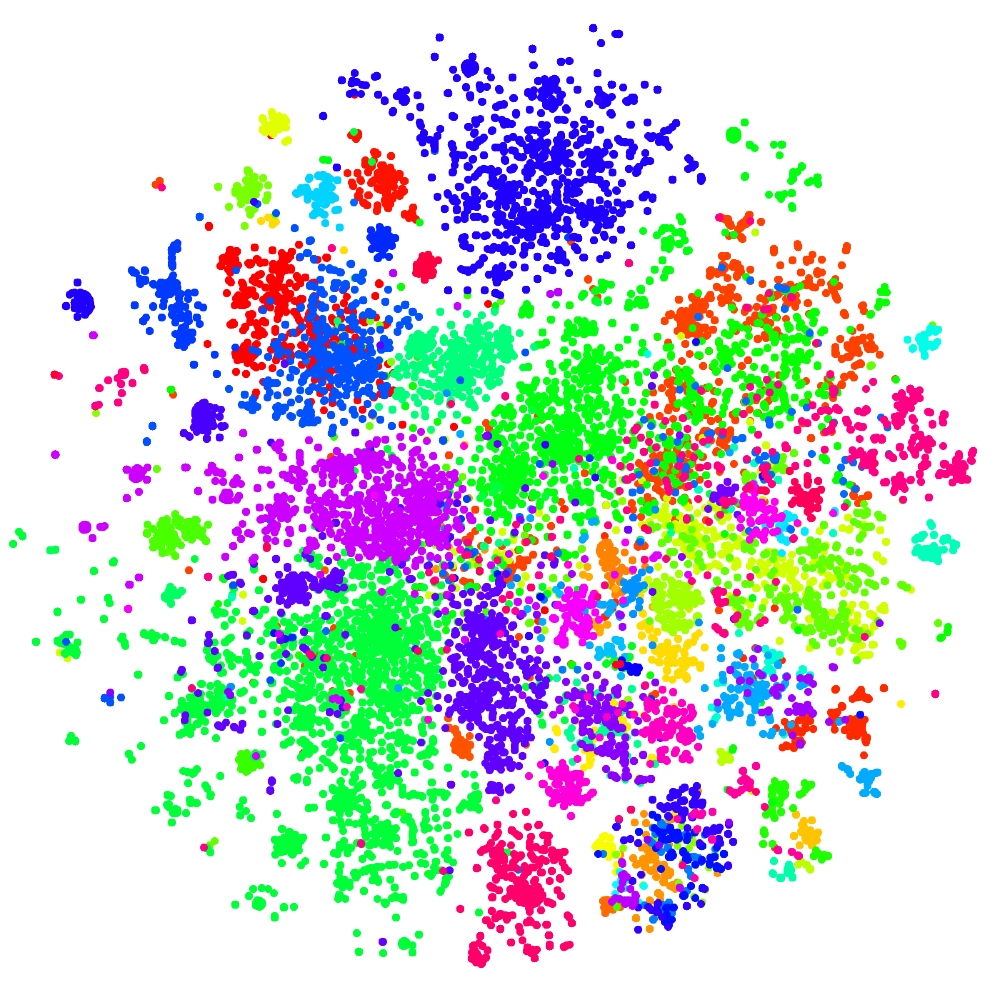}
        \end{minipage}
	}
	\subfigure[(Event) DEC]{
    	\begin{minipage}[t]{0.16\linewidth} 
		\centering \includegraphics[scale=0.08]{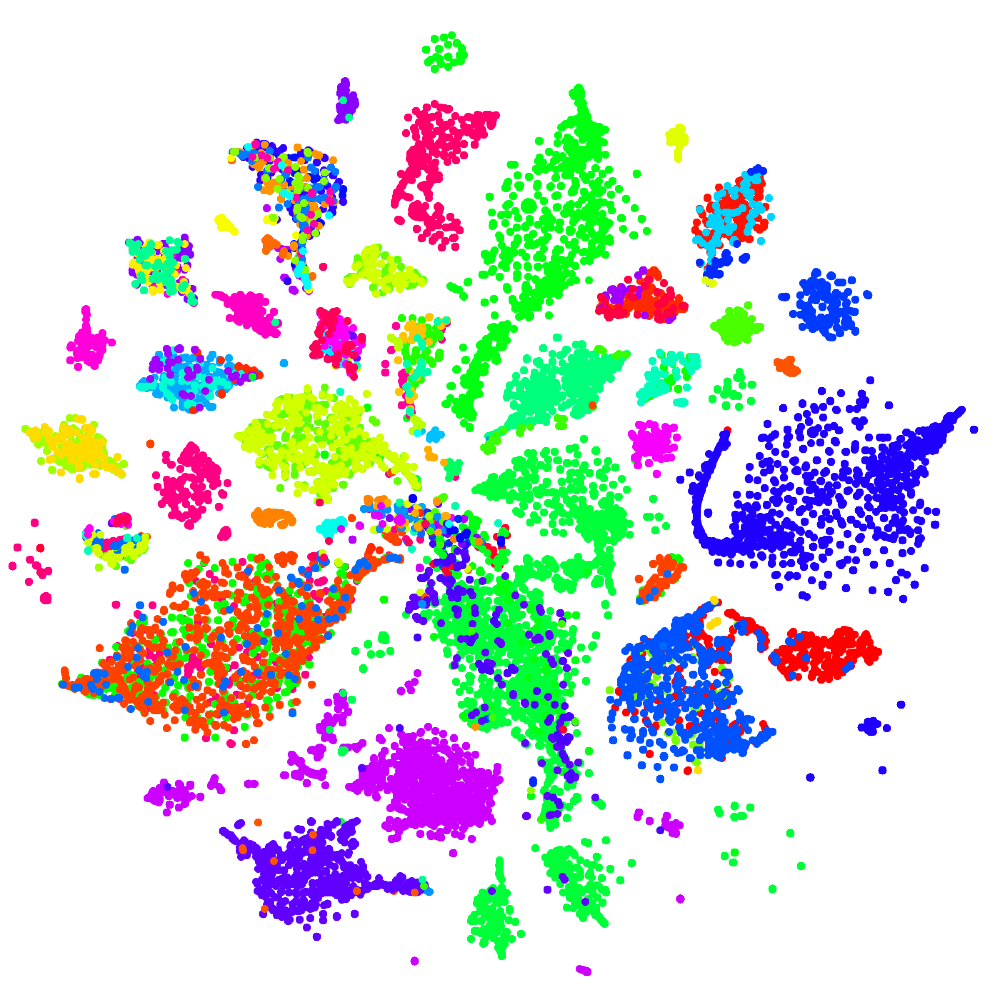}
        \end{minipage}
	}
	\subfigure[(Event) ARL]{
    	\begin{minipage}[t]{0.16\linewidth} 
		\centering \includegraphics[scale=0.08]{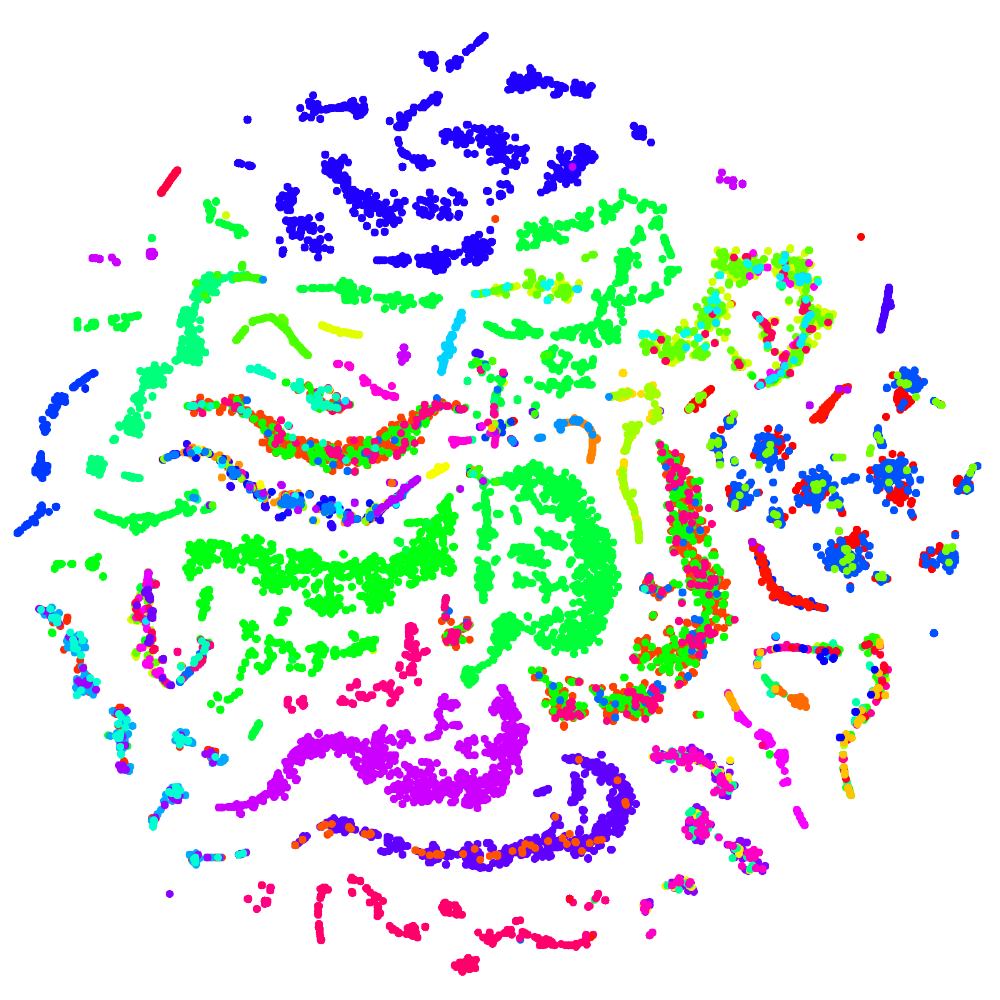}
        \end{minipage}
	}
	\subfigure[(Event) ARL-Adv]{
    	\begin{minipage}[t]{0.16\linewidth} 
		\centering \includegraphics[scale=0.08]{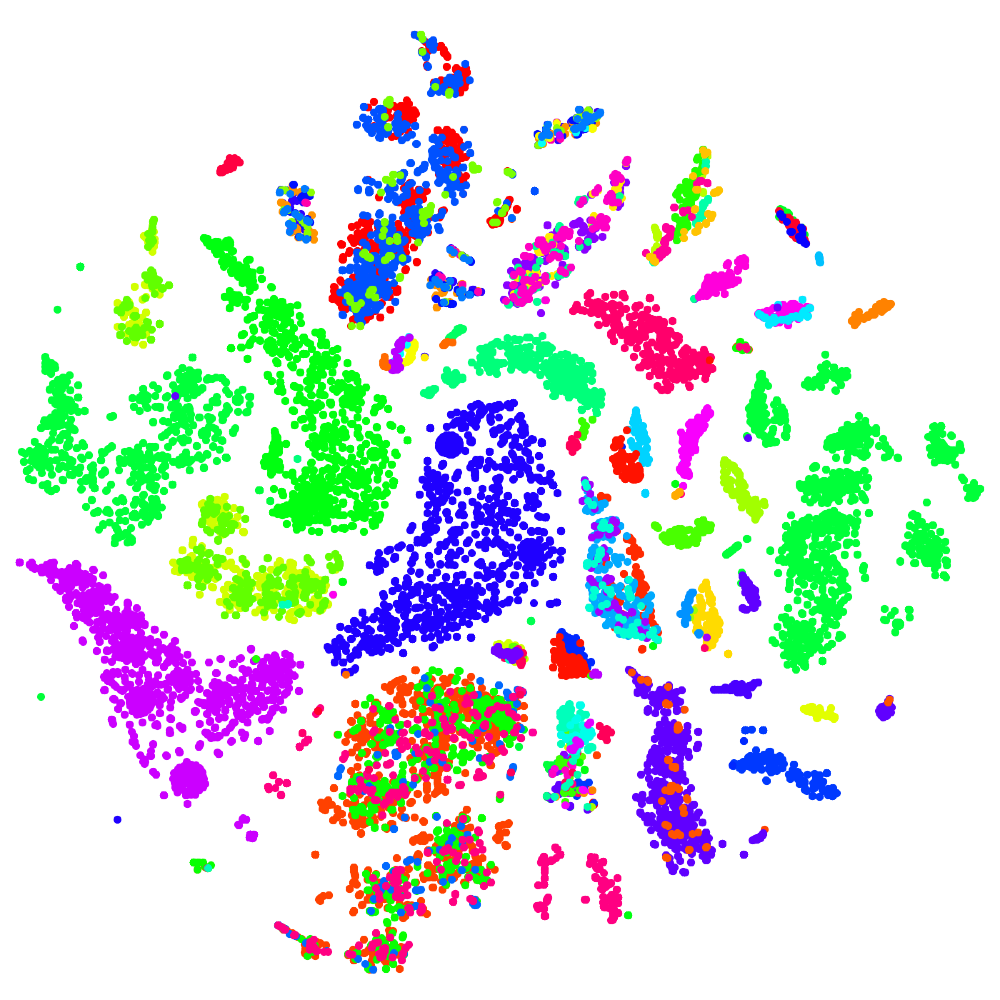}
        \end{minipage}
	}
    \subfigure[(SO) TF-IDF]{
    	\begin{minipage}[t]{0.16\linewidth} 
		\centering \includegraphics[scale=0.08]{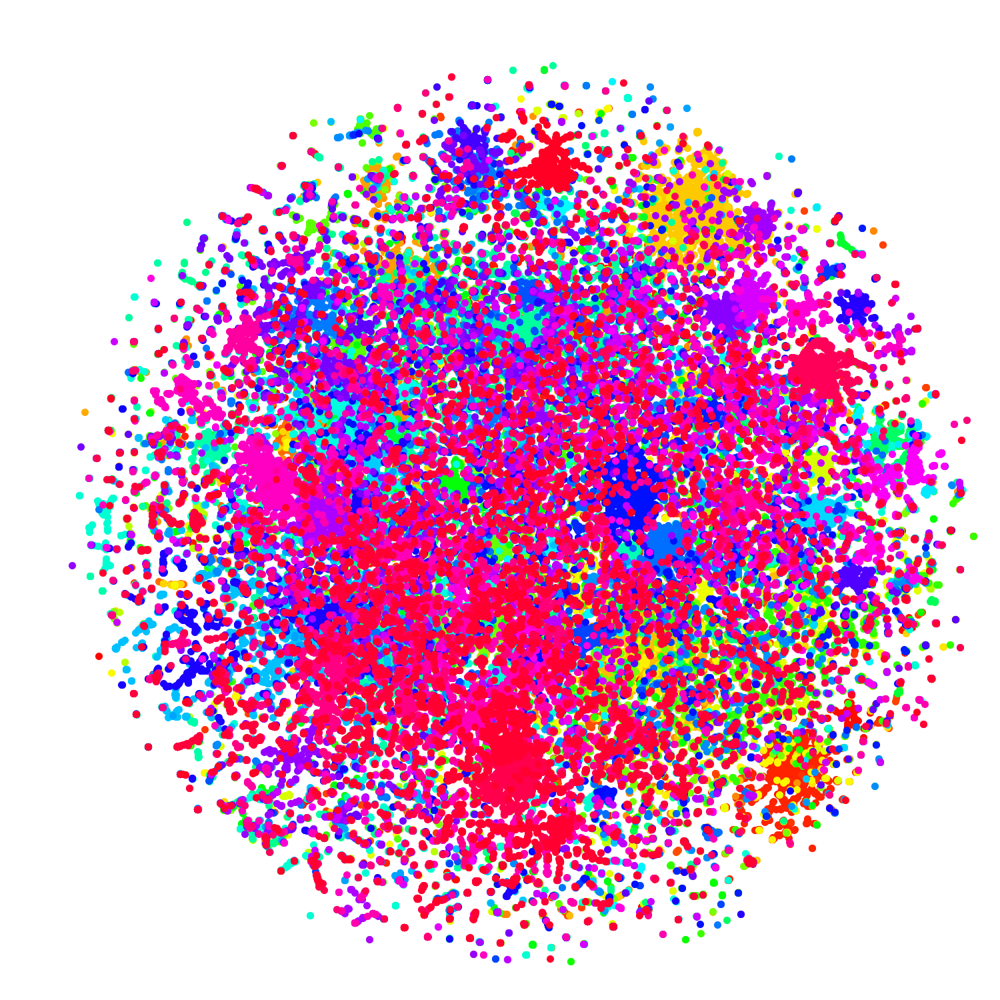}
        \end{minipage}
	}
	\subfigure[(SO) VaDE]{
    	\begin{minipage}[t]{0.16\linewidth} 
		\centering \includegraphics[scale=0.08]{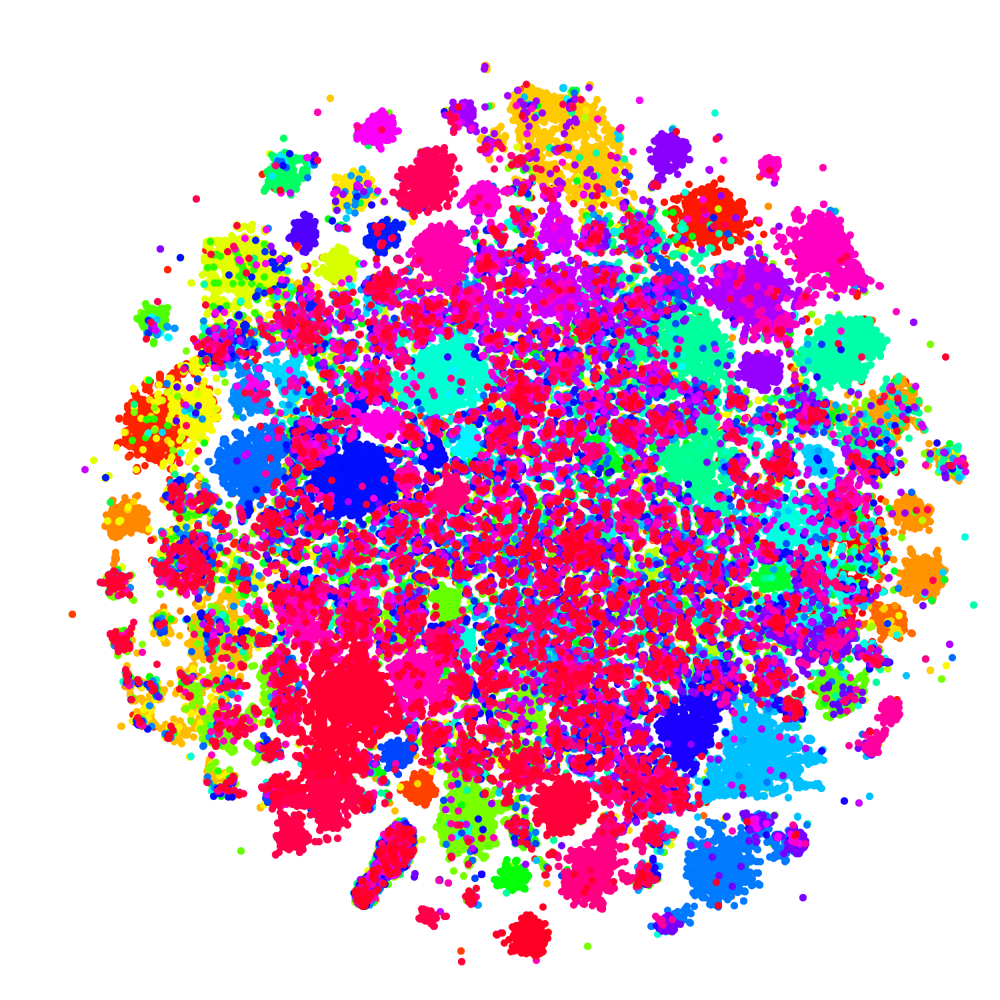}
        \end{minipage}
	}
	\subfigure[(SO) DEC]{
    	\begin{minipage}[t]{0.16\linewidth} 
		\centering \includegraphics[scale=0.08]{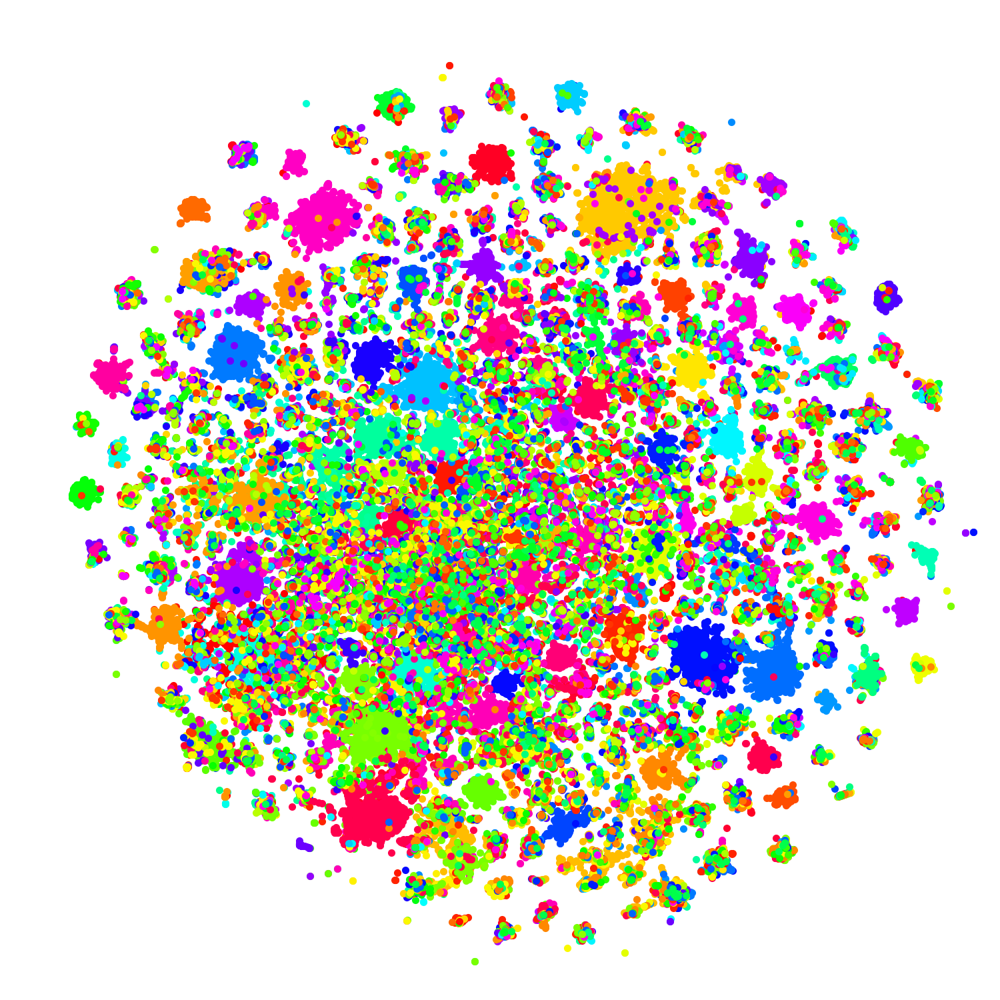}
        \end{minipage}
	}
	\subfigure[(SO) ARL]{
    	\begin{minipage}[t]{0.16\linewidth} 
		\centering \includegraphics[scale=0.08]{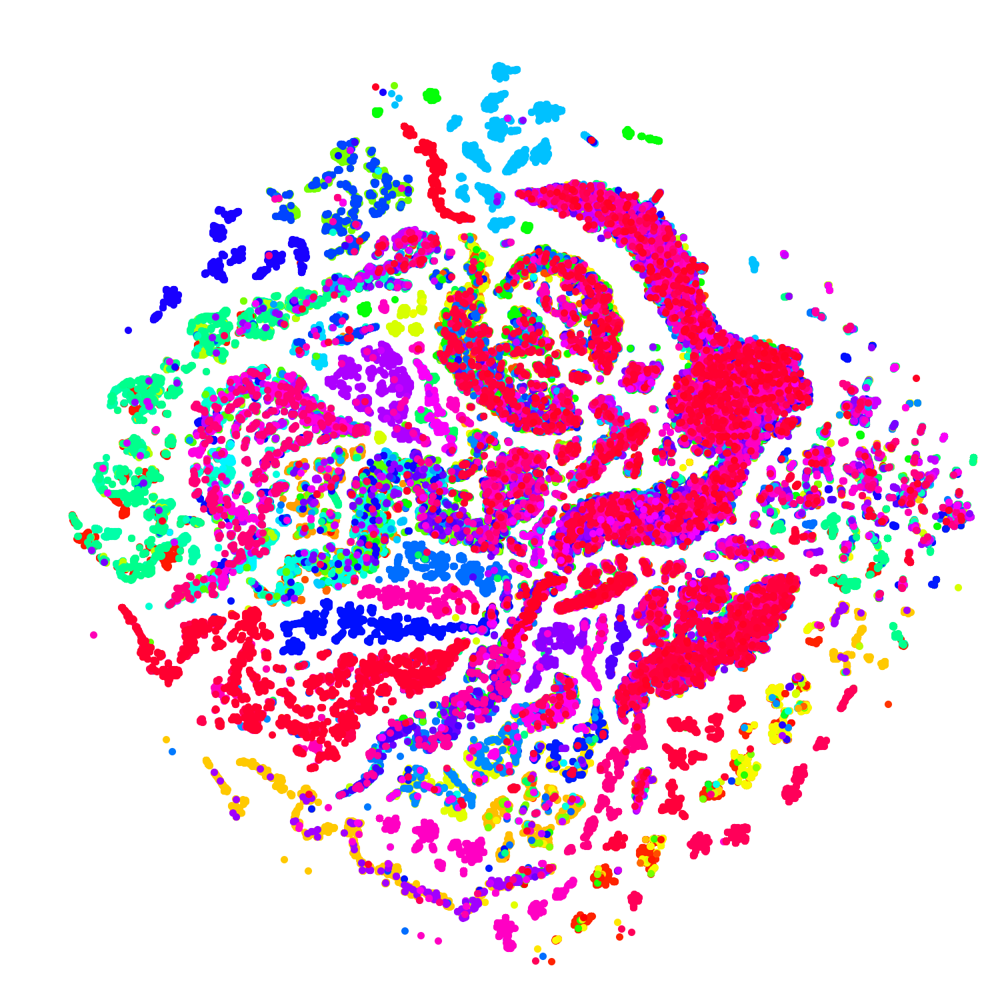}
        \end{minipage}
	}
	\subfigure[(SO) ARL-Adv]{
    	\begin{minipage}[t]{0.16\linewidth} 
		\centering \includegraphics[scale=0.08]{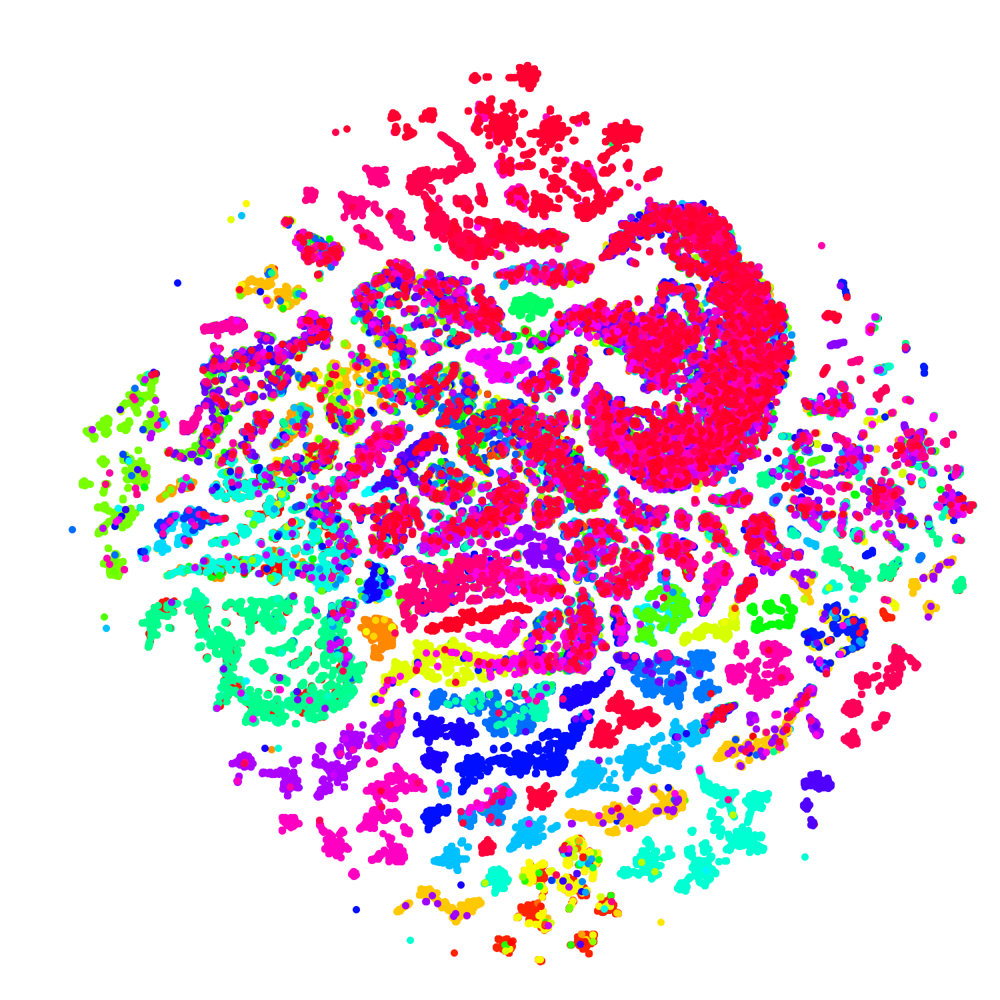}
        \end{minipage}
	}
	\vspace{-0.5em}
	\caption{
		Visualization of all short text representations in a 2-dimensional space using t-SNE.
		The quality of TF-IDF features, short text embeddings from VaDE, DEC, and the proposed models are investigated.
	}
	\label{fig:visualization}
\end{figure*}

\begin{figure*}
	\centering
	\includegraphics[width=.32\textwidth]{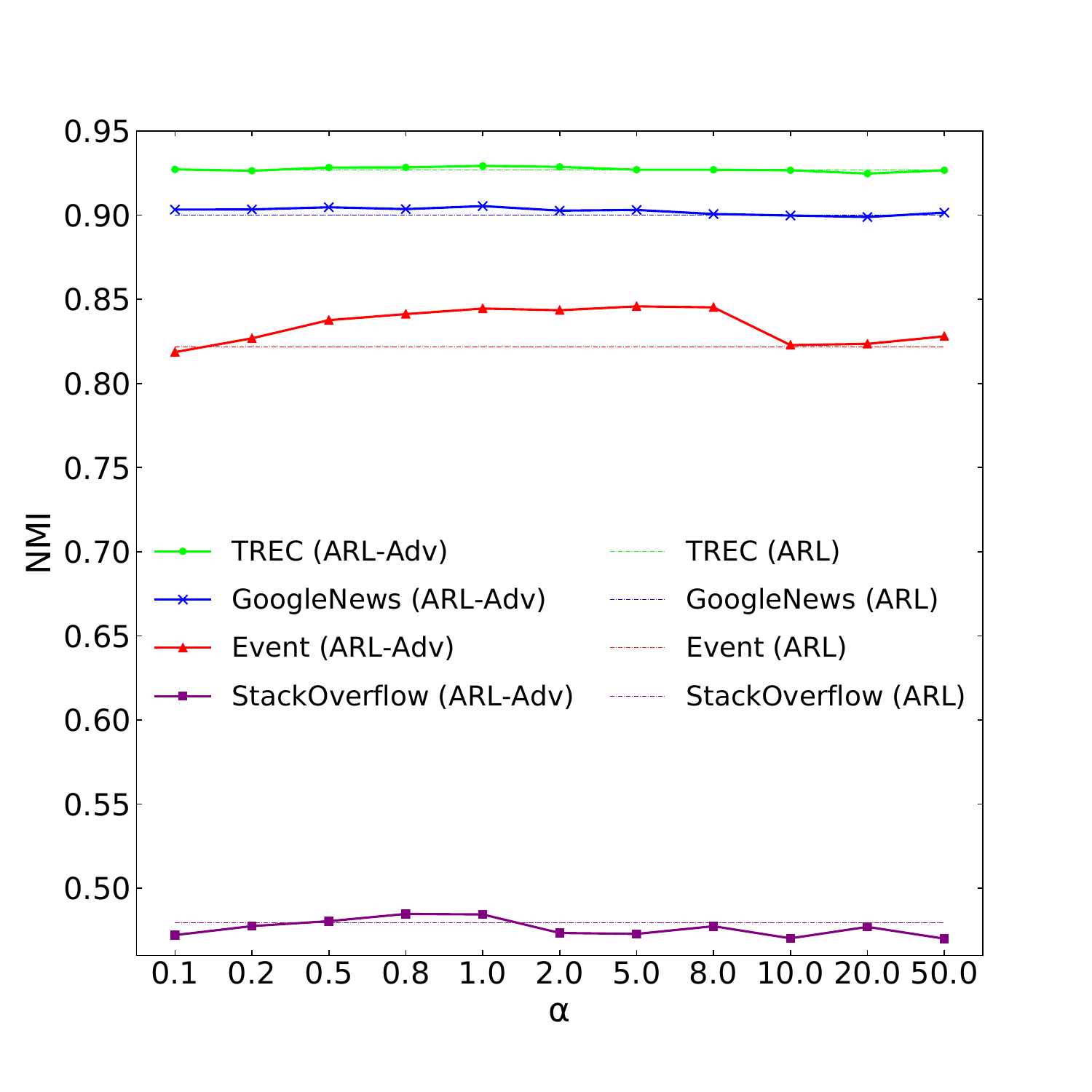}
	\includegraphics[width=.32\textwidth]{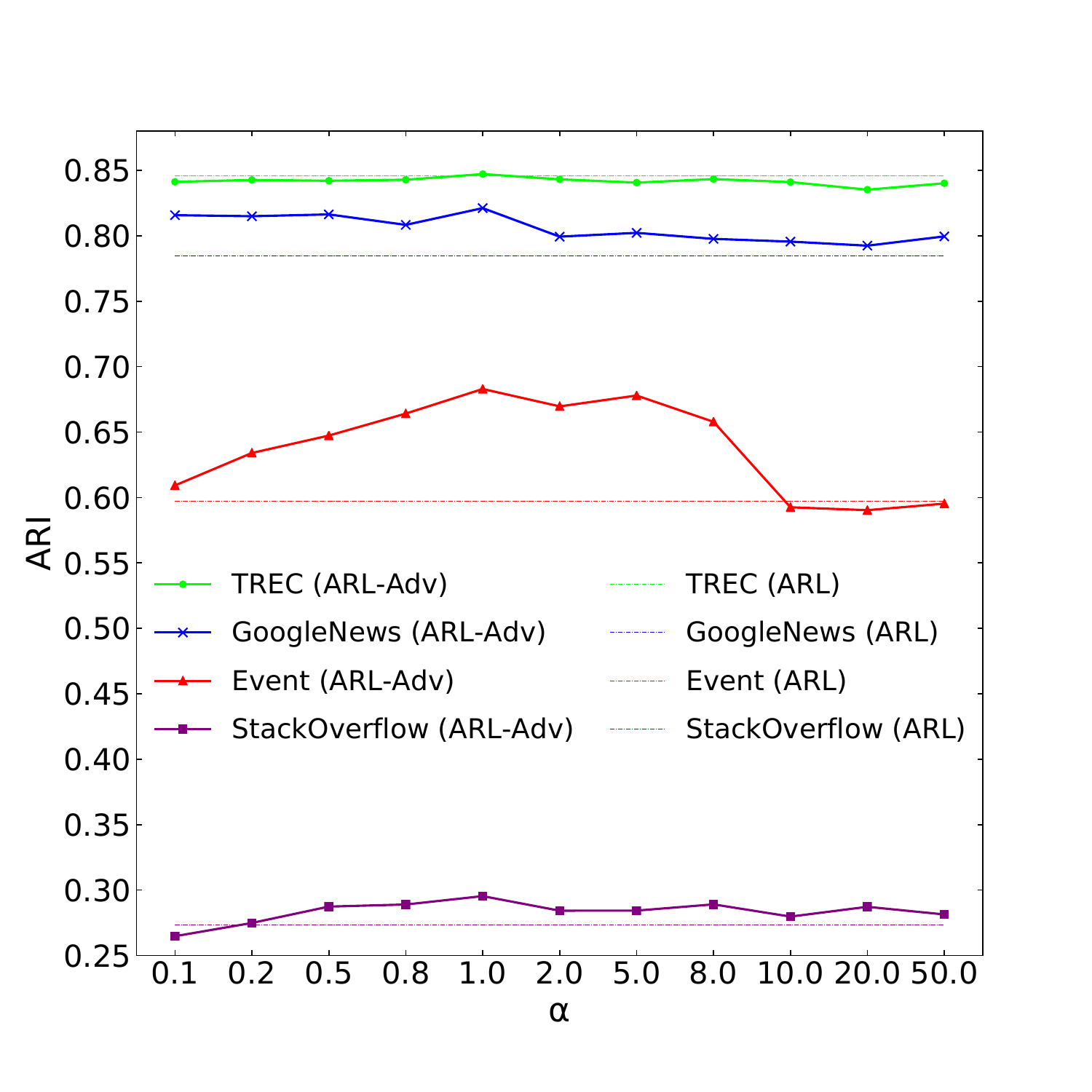}
	\includegraphics[width=.32\textwidth]{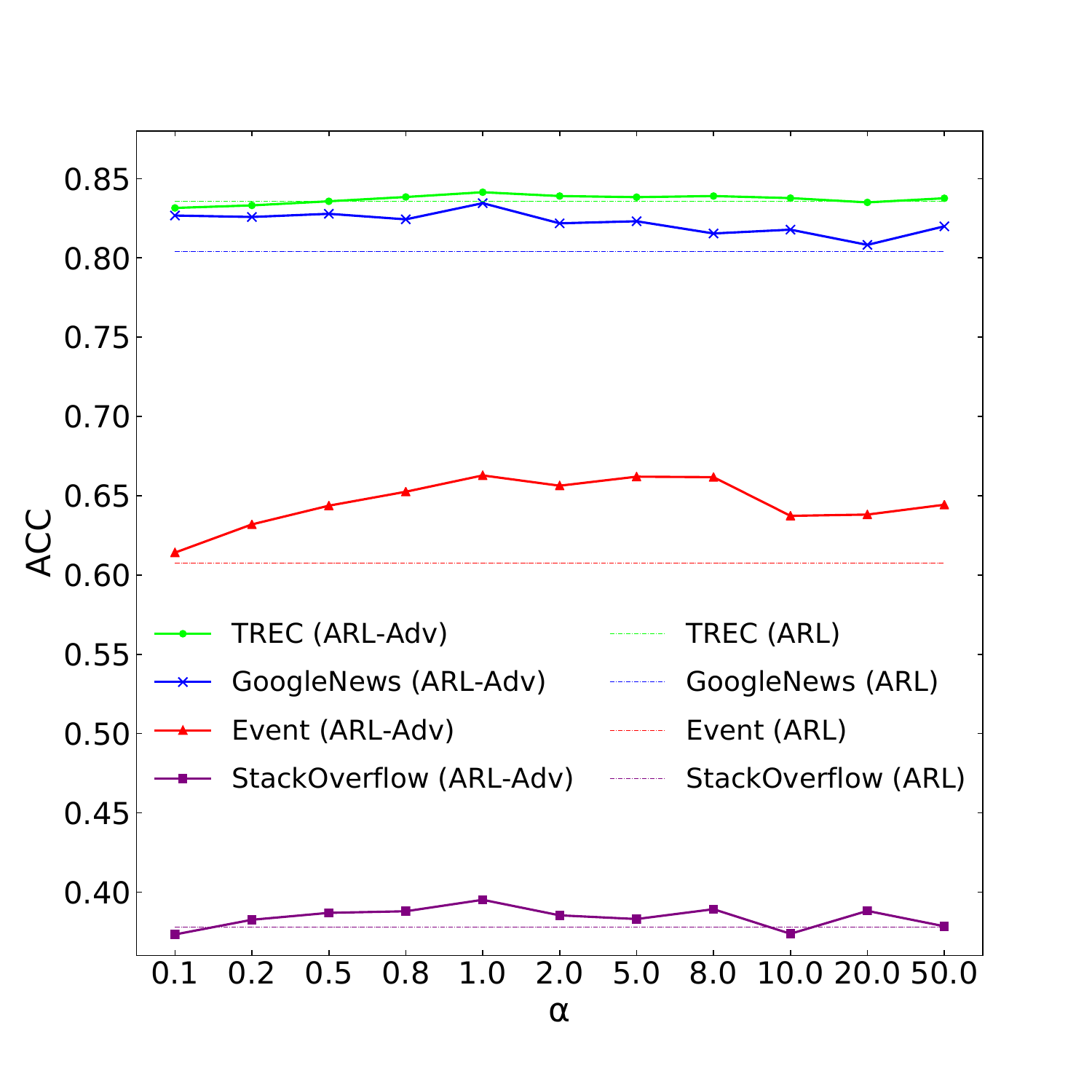}
	\caption{Influence of $\alpha$ on the model performance.}
	\label{fig:alpha_influence}
\end{figure*}

\begin{figure*}
	\centering
	\includegraphics[width=.32\textwidth]{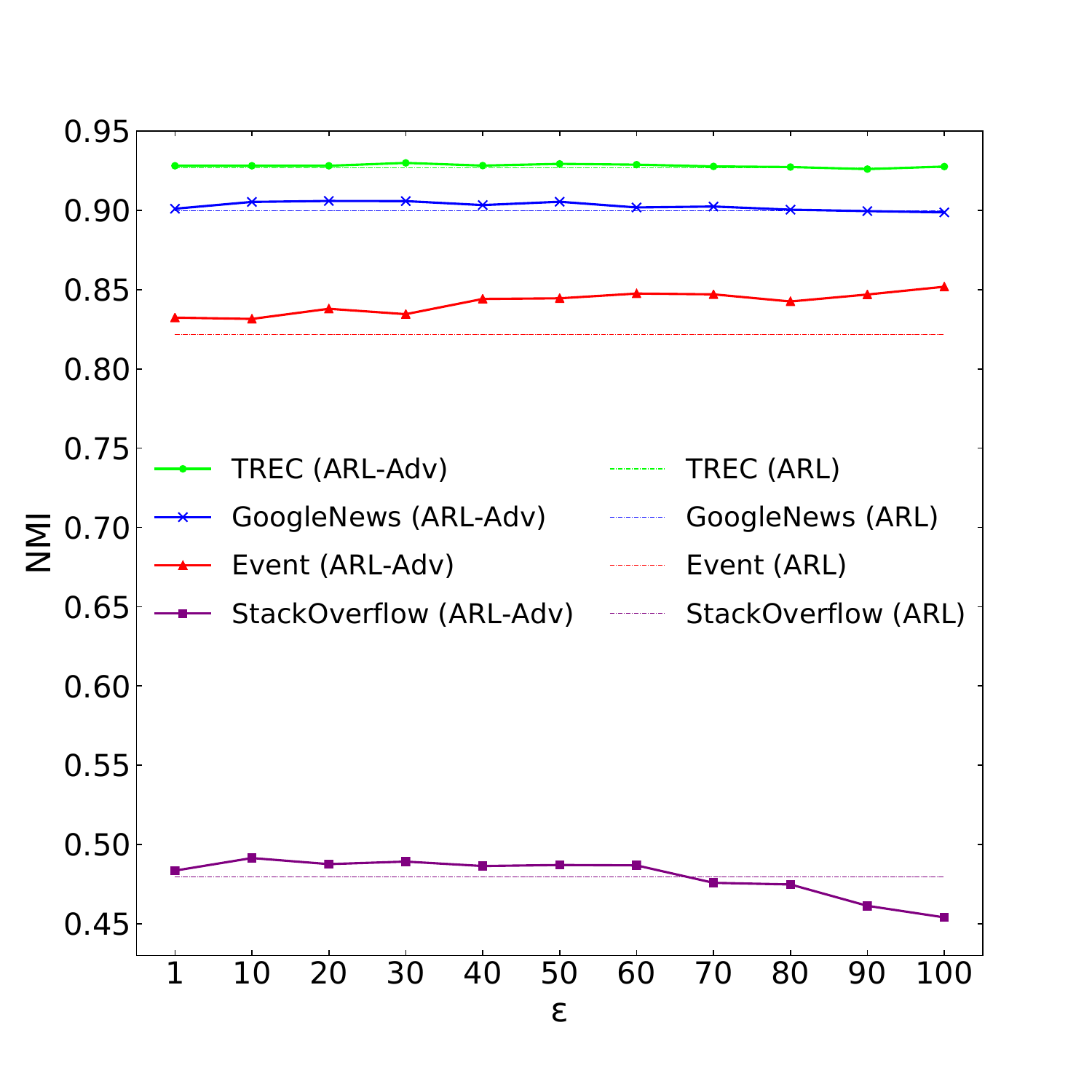}
	\includegraphics[width=.32\textwidth]{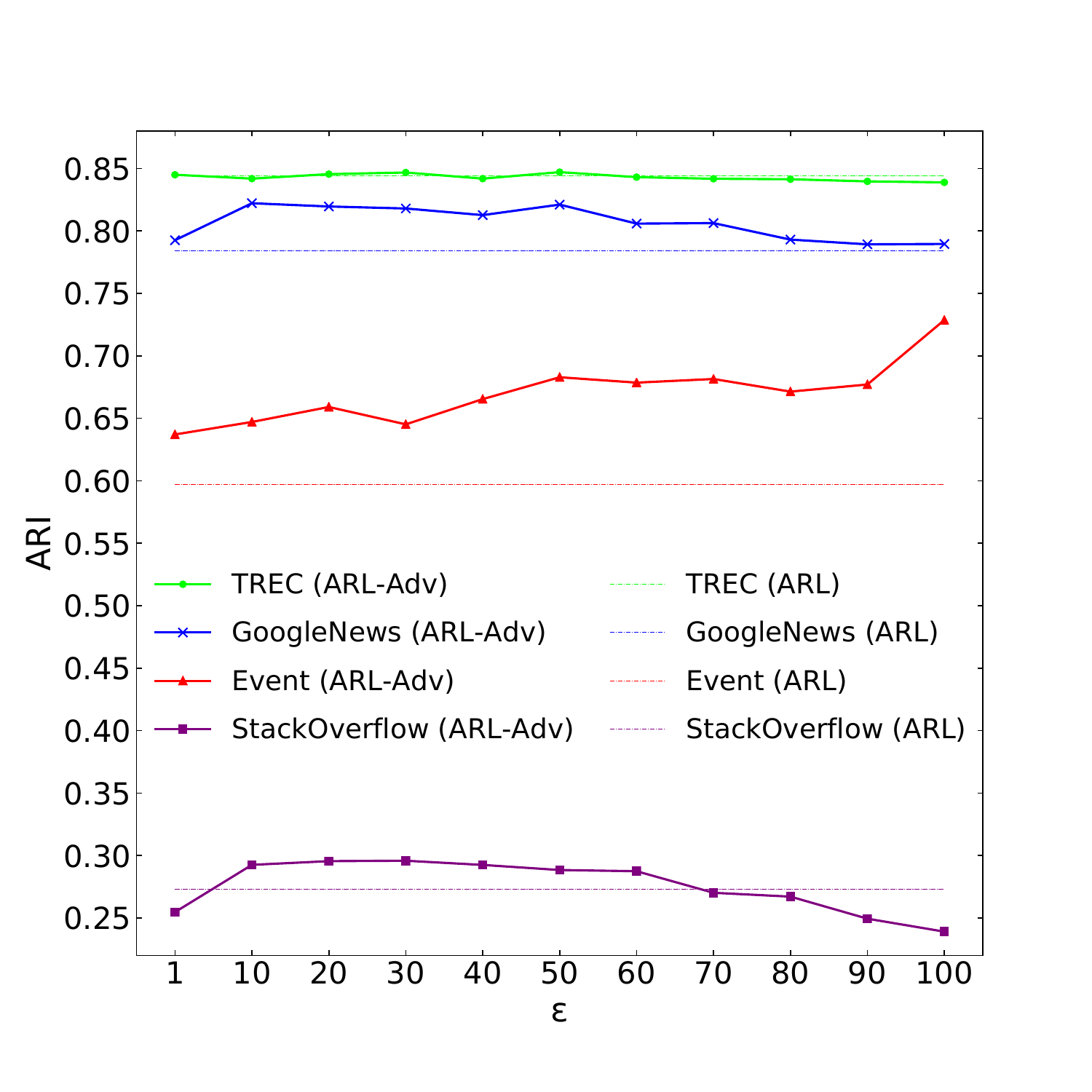}
	\includegraphics[width=.32\textwidth]{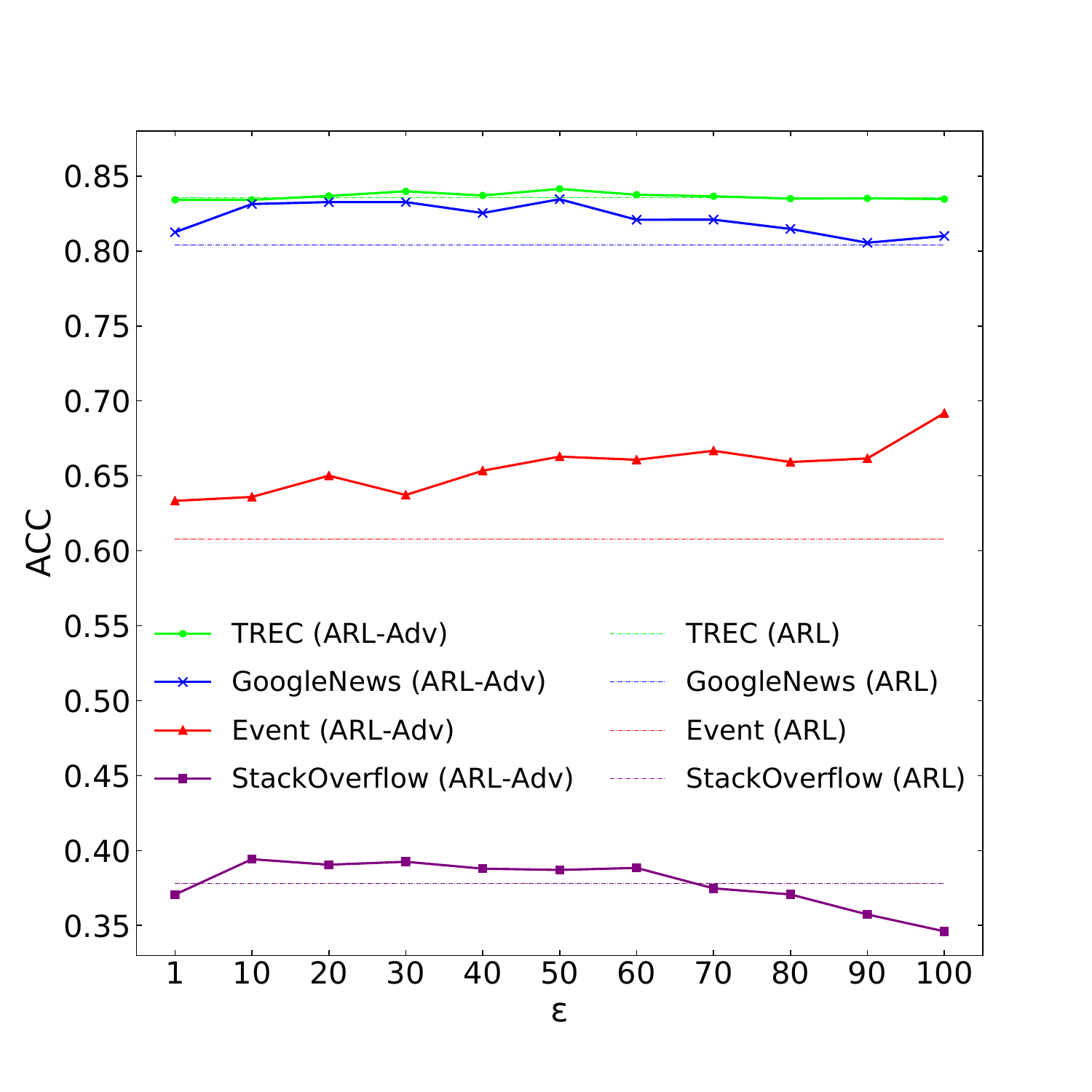}
	\caption{Influence of $\epsilon$ on the model performance.}
	\label{fig:eps_influence}
\end{figure*}

\subsection{Analysis of the Proposed Model (\textbf{\texttt{Q2}})}\noindent
Table~\ref{tbl:ablation} shows the results of the ablation study, so we can investigate the role of critical components in our proposed model.
By first comparing ARL-Adv(no train w) and ARL-Adv(no train c) with ARL-Adv, we can see their performance drops significantly.
This phenomenon shows the learning of word representations and cluster representations, accompanied by the automatic selection of clusters, is indeed indispensable.
Besides, ARL-Adv(no train c) behaves better than ARL-Adv(no train w), indicating the training of word embeddings might be more critical for the model.

The second part of Table~\ref{tbl:ablation} shows the contributions of the pairwise-ranking loss and pointwise loss.
For TREC, the two losses play similar roles, and removing either of them does not obviously damage the performance.
On the contrary, the pairwise ranking loss is more crucial for the clustering performance on the other three datasets, and adding the pointwise loss could strengthen short text clustering in most cases --- no improvement observed in StackOverflow.
Actually, we could add a hyper-parameter in Equation~\eqref{eq:j1} and~\eqref{eq:j2} to control the relative influence of the pointwise loss for a specific dataset.

The third part of Table~\ref{tbl:ablation} further explores the effect of adversarial cluster-level perturbations, by comparing it with cluster-level random perturbations (ARL-Random) and adversarial word-level perturbations (ARL-Adv(word)).
Based on the results, we observe that:
(1) ARL-Random obtains slightly worse results than ARL.
This reveals that simply adding random perturbations without learning may not bring more useful information to the model.
(2) ARL-Adv(word) performs marginally better than ARL in most cases, but not as well as ARL-Adv, especially for GoogleNews, Event, and StackOverflow. 
For example, compared to ARL-Adv(word), ARL-Adv gains the relative improvements of $2.8\%$, $4.8\%$, and $7.0\%$ evaluated by ACC for the three datasets, respectively.

\begin{table}[!t]
	\centering
	\setlength{\tabcolsep}{0.2pt}
	\caption{Representative topical words of TREC.}
	\label{tbl:vs_word_trec}
	\begin{small}\begin{tabular}{m{0.8cm}<{\centering}|m{4cm}<{\centering}|m{4cm}<{\centering}}
	    \toprule[1.3pt]
			Topic   &ARL-Adv		&GSDMM \\
			\hline
			1
			& card consolid credit debt loan unsecur advic settlement counsel creditor bailout
			& debt credit consolid card loan settlement relief bad negoti post unsecur combin lower \\
			\hline
			2
			& gifford gabriell buildup fluid congresswoman rehabilit icu recoveri brain hospit doctor
			& gifford gabriell rehab doctor recoveri brain news congresswoman fluid buildup \\
			\hline
			3
			& drone uav deliveri airspac southwest amazon chair requir commerci propos slice
			& drone amazon propos airspac deliveri commerci zone sky fli uav high-spe plan  \\
	    \bottomrule[1.3pt]
		\end{tabular}\end{small}
\end{table}

\begin{table}[!t]
	\centering
	\caption{Representative topical words of Event.}
	\setlength{\tabcolsep}{0.2pt}
	\begin{small}\begin{tabular}{m{0.8cm}<{\centering}|m{4cm}<{\centering}|m{4cm}<{\centering}}
    \toprule[1.3pt]
		Topic   &ARL-Adv		&GSDMM \\
		\hline
		1
		& daallo plane somalia land somali beachfront shabaab au hablod mogadishu airplan 
		& somalia plane land emerg explos forc daallo airlin injur make passeng hole blast  \\
		\hline
		2
		& plow plough bastilleday nice revel speed promenad driven franceattack truck
		& truck crowd attack prayfornic innoc heart shatter tie enjoy parad plow leader  \\
		\hline
		3
		& lahoreblast pakistanbomb lahor peshawar prayforlahor peshawarblast guess 
		& lahoreblast islamabad pakistan pm armi punjab govt oper blast lahor minist call  \\
    \bottomrule[1.3pt]
	\end{tabular}\end{small}
	\label{tbl:vs_word_event}
\end{table}

\subsection{Qualitative Studies (\textbf{\texttt{Q3}})}\noindent
To explore the representations generated by ALR-Adv and other baselines, we use t-SNE~\cite{Maaten08-JMLR} to visualize their short text embeddings in Figure~\ref{fig:visualization}, where one color denotes a cluster.
We find that ARL-Adv and ARL generate highly separable semantic clusters, compared with high-dimensional yet sparse TF-IDF based features and low-dimensional embeddings from VaDE and DEC.
This visualization explains the outstanding scores of ARL-Adv evaluated by the three metrics.
DEC shows better clustering visualization than the other two baselines.
However, the central areas in TREC and GoogleNews are somehow fuzzy, and clusters in Event and StackOverflow are not separated so well.
Even for humans, the clusters provided by ARL-Adv and ARL are easy to identify, exhibiting its potential as the basis or initialization for downstream applications that require insights into data manifold.
Besides, there exist non-negligible differences between the visualizations of ARL-Adv and ARL, showing robust adversarial training indeed affects cluster representations.
For example, embeddings provided by ARL-Adv tend to be more compact than those provided by ARL, which is more apparent for Event and StackOverflow.

We further evaluate the cluster-level quality by comparing the keywords obtained by ARL-Adv and GSDMM, presented in Table~\ref{tbl:vs_word_trec} and~\ref{tbl:vs_word_event}.
One way of regarding a topic as being detected is to judge if it is held by most short texts belonging to a cluster.
Hence, we can align clusters from different methods according to the detected topics.
STCC, VaDE, and DEC may not afford such information since they do not directly model the correlations between words and clusters.
While GSDMM chooses the words based on the learned word distributions for different topics, ARL-Adv selects representative words that have larger cosine similarity values with the learned cluster representations.
The words that without identifiable meanings are discarded and the top listed words from the remaining are chosen.

Both ARL-Adv and GSDMM share lots of keywords in common, showing that they detect similar topics and reach consensus to a certain degree on words that better reflect the topics.
Taking topic 2 of TREC as an example, words like $brain$, $fluid$, and $recoveri$ imply the case of a cured illness, related to the entity \textit{gifford} or $gabriell$.
However, GSDMM only identifies $doctor$, while ARL-Adv further points out $hospit$ and $icu$.
The two words are even not ranked higher in the word lists of GSDMM, yet they are indeed highly correlated to the topic.
For Event, we can observe a similar phenomenon that overlapped keywords still exist between the two methods, confirming their capability of providing an overview of a cluster.
As revealed in topic 1 of Event, while both of them succeed to identify $dallo$, $somalia$, and other words, ARL-Adv goes further and extracts highly correlated entities like $beachfront$ and $shabaab$.
These evidences consolidate our viewpoint that ARL-Adv can capture highly correlated keywords, which tend to be neglected by conventional topic models like GSDMM.
Hence our model is welcomed by better semantic quality at the cluster level.

\begin{figure*}[!t]
	\centering
	\includegraphics[width=.29\textwidth]{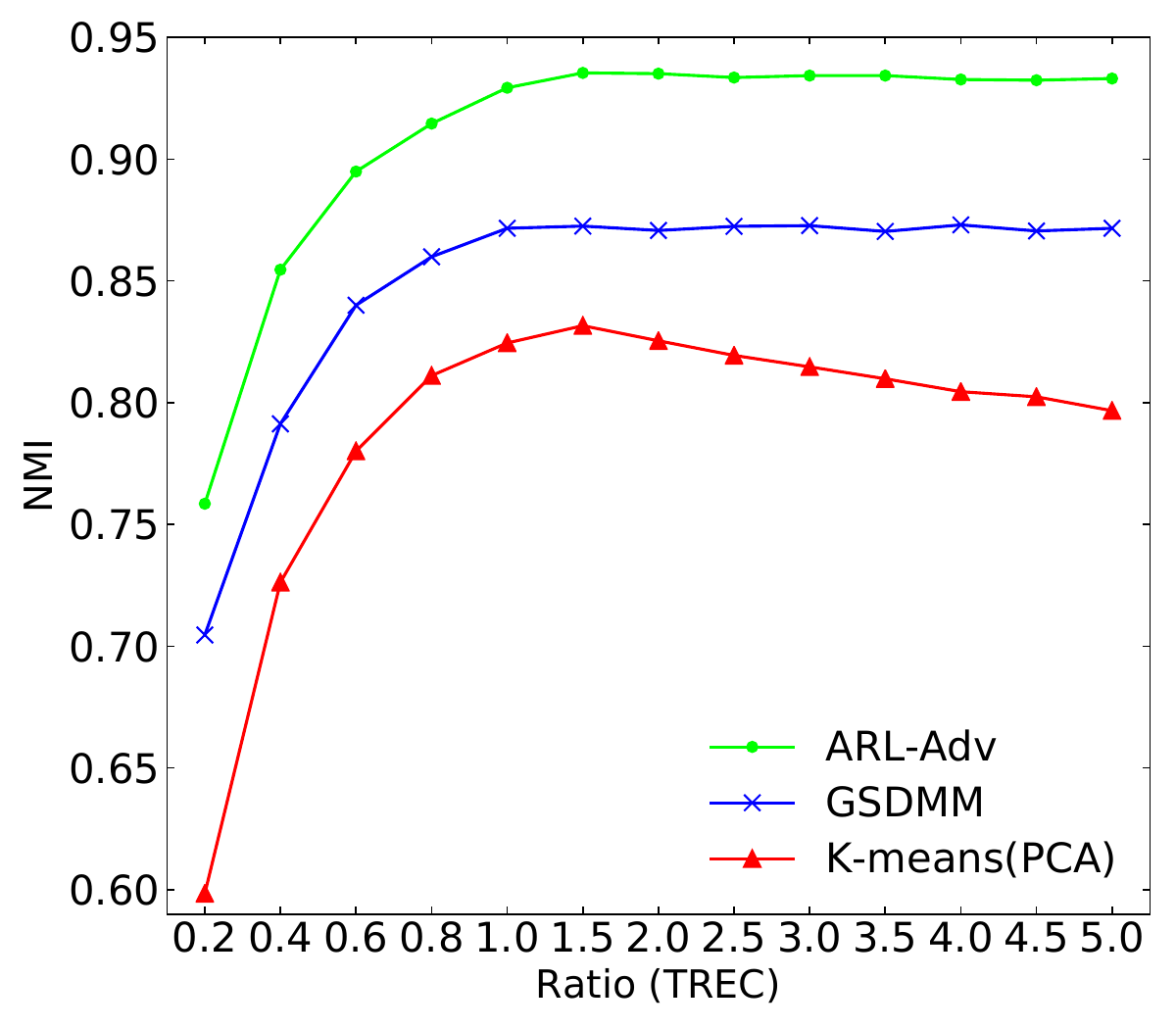}\hspace{30pt}
	\includegraphics[width=.29\textwidth]{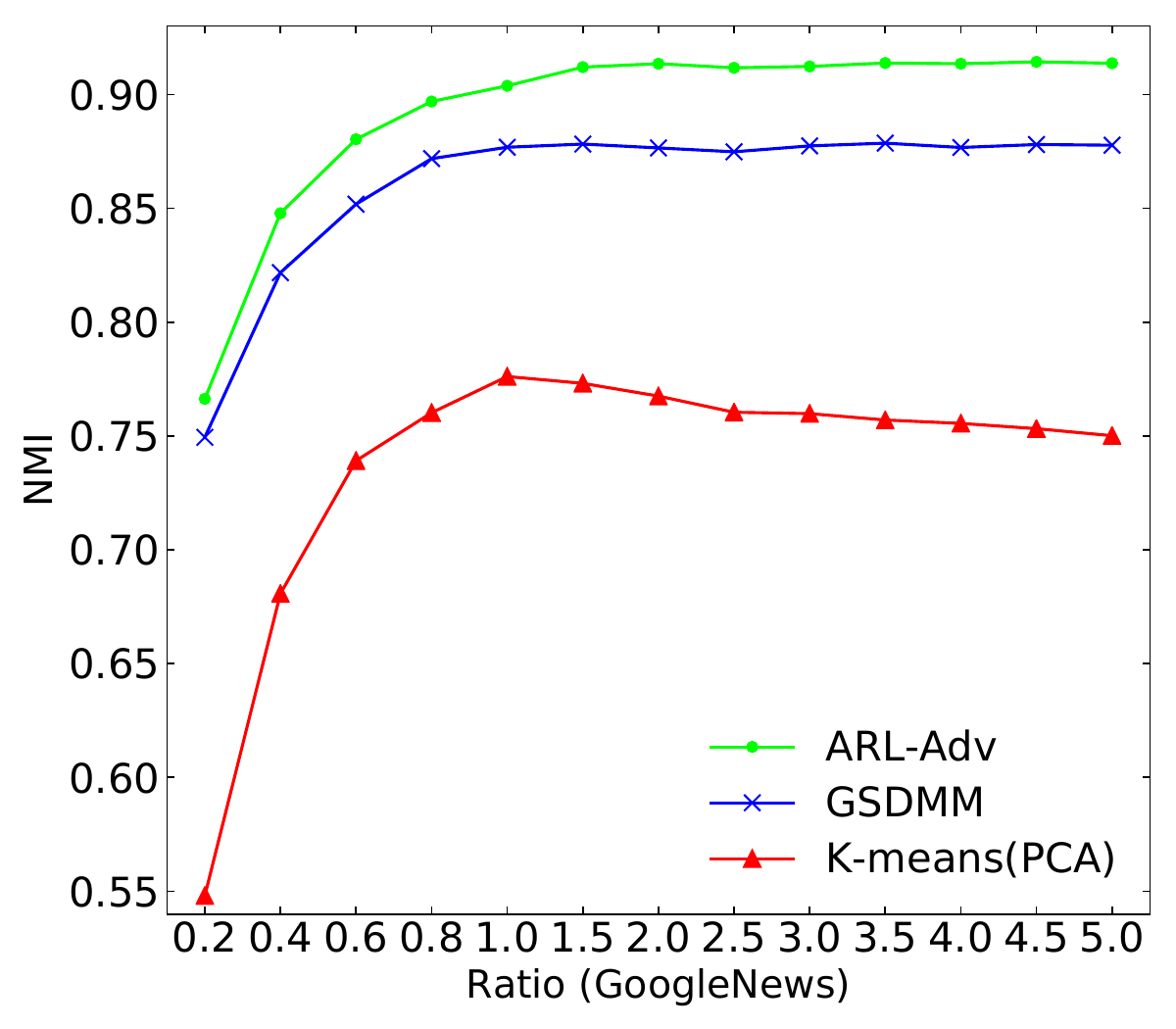}\hspace{30pt}
	\includegraphics[width=.29\textwidth]{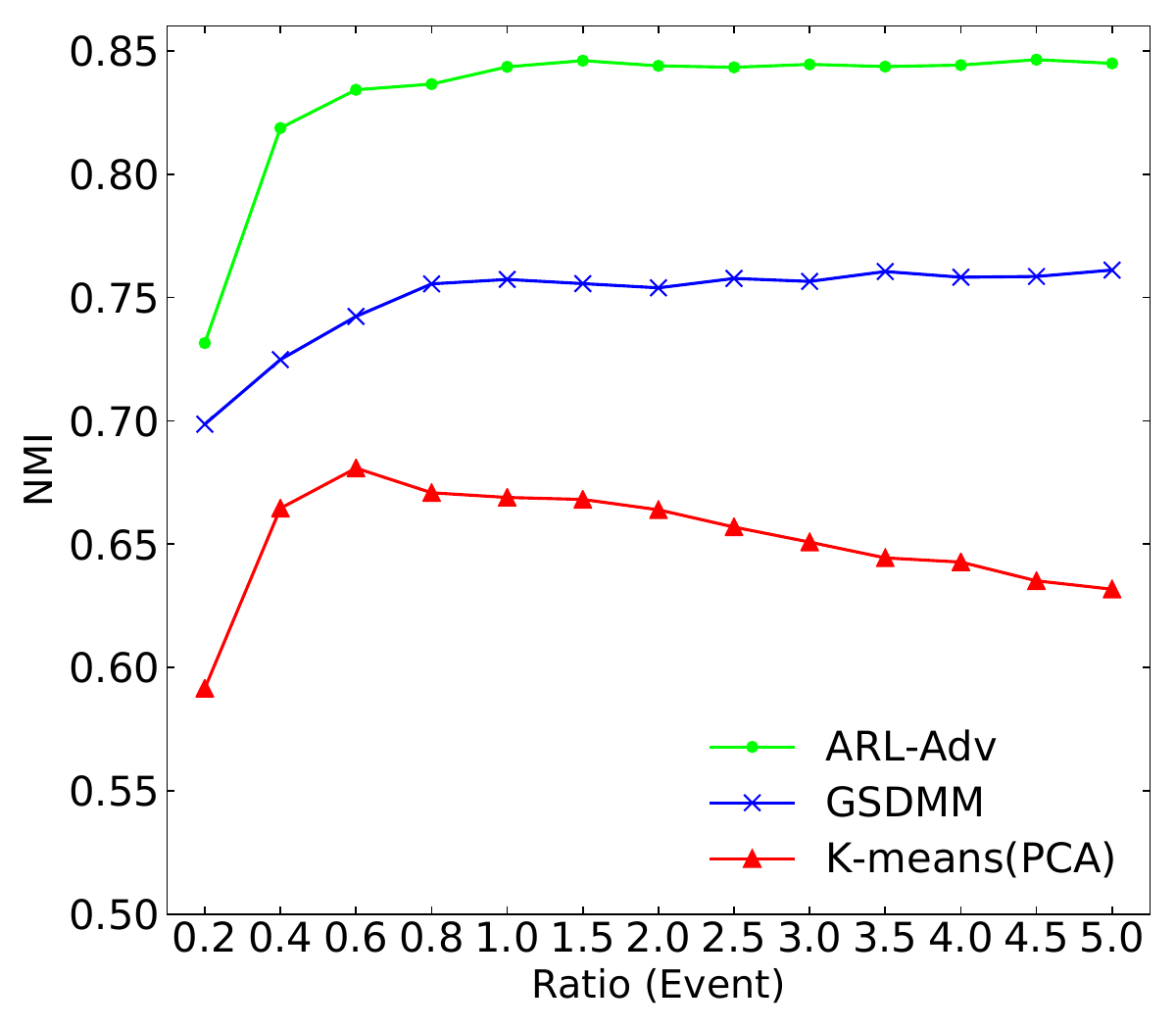}
	\caption{Influence of the cluster number. The ratio of intended cluster number to the original topic number is used. }
	\label{fig:clu_nums}
\end{figure*}

\subsection{Hyper-parameter Analysis}\label{sec:hyperparameter}
\subsubsection{Effect of $\alpha$}
As described in Equation~\eqref{eq:opt-obj}, $\alpha$ adjusts the strength of $J_2$ and thus constrains the contribution of adversarial perturbations to the final loss function.
Figure~\ref{fig:alpha_influence} shows how the clustering performance of ARL-Adv alters along with the selection of $\alpha$, which helps us to understand the role of loss $J_2$.
Note that $\epsilon$ is set to the default value $50$ as shown in Table~\ref{tbl:hyper-setting}.
The figures show that a wide range of $\alpha$, from $0.2$ to $8$, can already provide a significant gain in most cases among the datasets and metrics.
Moreover, selecting $\alpha$ around $1.0$ achieves better results than most values in TREC, GoogleNews, and StackOverflow.
Thus reaching a balance between the loss function $J_1(\itbold{E}, \itbold{C})$ and its adversarial version $J_2(\itbold{E}, \itbold{C}+\Delta^C)$ is a preferable choice in ARL-Adv and we fix $\alpha$=$1.0$ throughout the experiments.
In summary, ARL-Adv relies, but not completely, on robust adversarial learning, showing that it appropriately and effectively takes the advantage of perturbations.

\subsubsection{Effect of $\epsilon$}
Figure~\ref{fig:eps_influence} presents the performance of ARL-Adv when varying $\epsilon$.
As defined in Equation~\eqref{eq:update-M}, $\epsilon$ constrains the norm of adversarial perturbations on cluster representations.
The larger $\epsilon$ is, the larger the adversarial perturbations will be during the training process.
Under the setting of $\alpha$ discussed above, we select some values for $\epsilon$ in the range from $1$ to $100$, and show how it affects the clustering performance.
We can infer from the figures that positive impact can be achieved locally when $\epsilon$ falls in the range [10, 70] in most cases.
Noticeably, while increasing $\epsilon$ based on the chosen range indeed shows a growing tendency for Event, it does not apply to the other three datasets.
Thus, we do not alter $\epsilon$ to larger values since it may not reflect the universal property of ARL-Adv.
Besides, varying $\epsilon$ below $1$ does not seem to be helpful, so we do not list the results here due to the limited space.
Consequently, we can summarize that the restriction imposed on the norm of $\Delta^C$ should lie in a proper scope to get reasonable results.

\begin{figure}[!t]
	\centering
	\includegraphics[width=.3\textwidth]{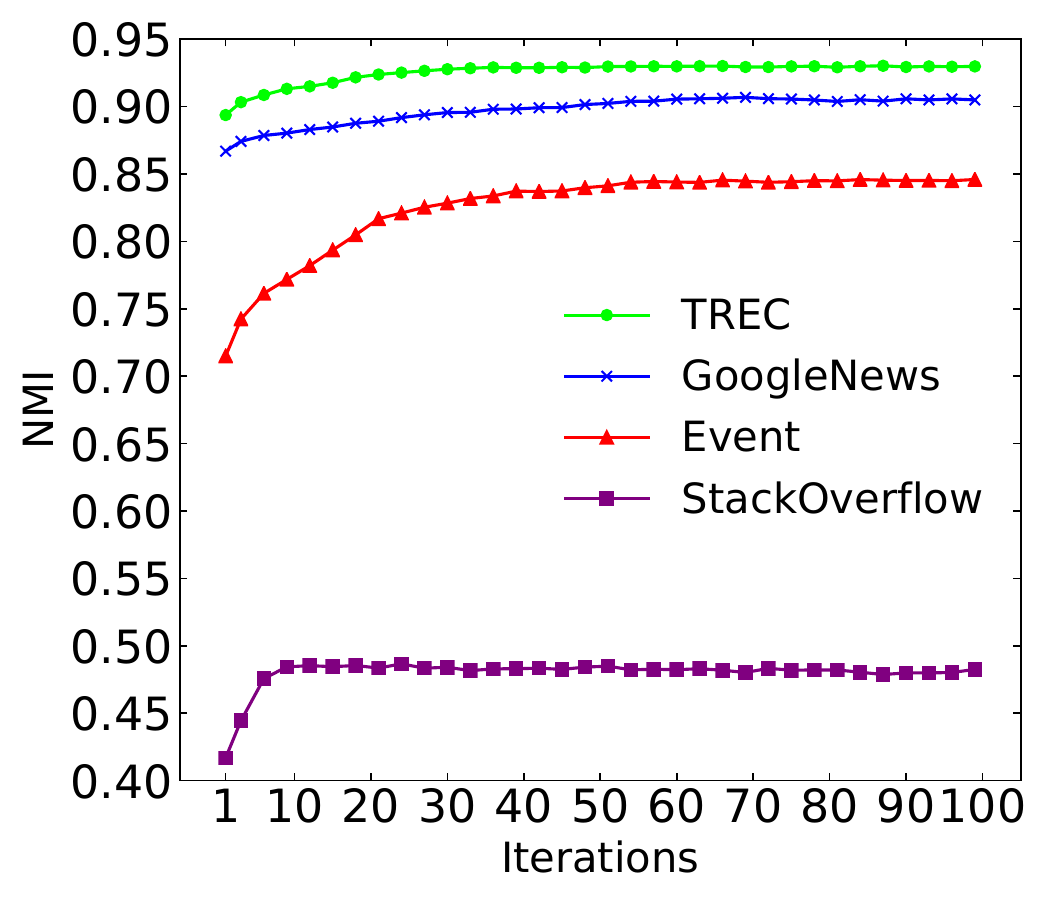}
	\caption{Performance over iterations.}
	\label{fig:iterations}
\end{figure}

\subsubsection{Effect of Cluster Number}
Figure~\ref{fig:clu_nums} shows how ARL-Adv adapts to the change of cluster number on the first three datasets w.r.t. NMI.
The ratio of the predetermined cluster number to the original topic number is used for convenience.
The initialization of cluster representations is the same as described in Section~\ref{subsubsec:arl_variants}.
We choose to compare ARL-Adv against GSDMM, which is one of the most competitive baseline, and K-means(PCA), which offers insights from the perspective of spectral clustering.
For all the considered methods, a dramatic increase is found when the ratio goes from $0.2$ to $0.4$.
This phenomenon is intuitive since the models are forced to merge texts from different topics into several clusters, which will almost inevitably lead to worse clustering results w.r.t. NMI.
After that, the evaluation scores continue to grow, and gradually get saturated.
K-means(PCA) fails to keep its performance when the ratio gets larger.
Therefore, spectral-based clustering algorithms might lack flexibility in the context of short text clustering, and turn out to heavily rely on human expertise in the datasets.
On the other hand, while increasing the ratio does not help too much for both ARL-Adv and GSDMM, ARL-Adv outperforms GSDMM in all situations by a large margin. 
This might be attributed to that they tend to keep their assignment strategies still even when the cluster counts are more than sufficient.
As such, ARL-Adv is not only effective, but also adaptive and robust enough to determine the proper mapping for texts when facing more options.

\subsubsection{Effect of Iterations}
We show the performance of the proposed ARL-Adv over iterations in Figure~\ref{fig:iterations}.
The curves are averaged over $10$ runs and we choose a few points at each interval for clarity.
Due to the effect of adversarial perturbations, ARL-Adv experiences a relatively moderate growth in its performance.
After $50$ or more iterations, ARL-Adv starts to attain a stable state on all metrics, sharing a good convergence property.

\subsection{Time Complexity Analysis}\label{subsec:time-analysis}
In this section, we move to compare the training time efficiencies for the adopted representation learning approaches on a single GPU.
Probabilistic topic models like GSDMM are not included since their Gibbs sampling based optimization implementations do not leverage the computational power of GPU well.
We also do not report the efficiencies of K-means algorithms since they demand multiple processors to speed up the solution of assigning each short text to its closest center.
To make fair comparisons, we set the batch size to 64 uniformity when testing the time costs of the methods to be compared.

\begin{table}[!t]
	\centering
	\caption{Training time costs (time unit: s).}
	\label{tbl:time-efficiency}
	\begin{tabular}{c|c|c|c|c|c|c}
	    \toprule[1.3pt]
		Data & ClusterGAN & DEC & VaDE & STC & ARL & ARL-Adv \\
		\hline 
        TREC & 3.7 & 1.1 & 2.2 & 1.1 & 1.7 & 2.1 \\
        SO & 42.9 & 30.2 & 40.1 & 27.3 & 24.5 & 27.6 \\
	    \bottomrule[1.3pt]
	\end{tabular}
\end{table}

Table~\ref{tbl:time-efficiency} shows the time costs needed for each training iteration in average, from which we have the following three key observations.
First, the computational efficiencies of the proposed models are comparable with the existing approaches.
Second, adopting robust adversarial training does not additionally incur a very heavy burden.
Third, an approximate linear increase in computational complexity of ARL and ARL-Adv is observed by the time cost and data size comparisons between TREC and StackOverflow (SO).
Since both ARL and ARL-Adv require the same or less training iterations --- usually less than 60 as shown in Figure~\ref{fig:iterations} --- for convergence compared to the other approaches through the experiments, we could conclude that their computational complexities are at a good level.
The results meet expectation since the proposed models have simple architectures and the model parameters only involve word embeddings, cluster representations, and adversarial perturbations of clusters.

\subsection{Application on Cluster-based Retrieval}\label{subsec:cluster-based retrieval}
To verify the effect of short-text clustering models on the downstream applications, we choose the cluster-based retrieval task~\cite{LiuC04} as a specific scenario for testing.
The task assumes that if a document is relevant to a given query, then the document cluster it belongs to should be related to the query as well.
Specifically, we select CBDM~\cite{LiuC04}, a variant of language models with cluster smoothing, as the retrieval model to validate the contribution of text clusters to retrieval performance.
The clusters we choose are produced by ARL-Adv and some representative baselines, including K-means(PCA), GSDMM, and DEC.

\begin{table}[!t]
\centering
\caption{Results of cluster-based retrieval. Noting that ``---'' denotes the original language model for retrieval without using clusters.}\label{tbl:retrieval}
\resizebox{.8\linewidth}{!}{
\begin{tabular}{c|cc} 
\toprule[1.2pt]
\multirow{2}*{Clustering Method}& \multicolumn{2}{c}{Precision@50} \\\cline{2-3}
&Event & StackOverflow \\\midrule[1.1pt]
--- & 0.6333 & 0.8561 \\\hline
K-means(PCA) & 0.6357 & 0.8550 \\\hline
GSDMM & 0.6423 & 0.8542\\\hline
DEC & 0.6426 & 0.8571 \\\hline
\textbf{ARL-Adv} & \textbf{0.6472} & \textbf{0.8647} \\\bottomrule[1.2pt]
\end{tabular}
}
\end{table}

Table~\ref{tbl:retrieval} presents the retrieval results on the Event and StackOverflow datasets evaluated by Precision@50, from which we observe that: (1) Using the clusters generated by our model ARL-Adv indeed boosts the performance of the downstream application, i.e., cluster-based retrieval.
(2) The gains brought by ARL-Adv are the largest among the chosen clustering methods.

\section{Conclusion}\label{sec:conclude}
In this paper, we have developed the novel clustering model ARL-Adv. 
It fuses the short text representation learning and clustering in a unified model through the proposed cluster-level attention.
Adversarial perturbations are further added to cluster representations, enhancing the robustness and effectiveness of model training through a minimax game.
Extensive studies on the four real-life datasets show that ARL-Adv can achieve superior performance, even compared to the state-of-the-art methods for short text clustering and the recently developed deep learning based clustering models.
Further analysis of ARL-Adv is performed to explain the contributions of its components.


%\bibliographystyle{IEEEtran}
\bibliographystyle{abbrv}
\bibliography{references}

%\vspace{-1.em}

%\newpage


%\vfill


\end{document}